%% file: main.tex
\documentclass[letterpaper]{article} 
\usepackage{aaai2026}  
\usepackage{times}  
\usepackage{helvet}  
\usepackage{courier}  
\usepackage[hyphens]{url}  
\usepackage{graphicx} 
\usepackage{natbib}  
\usepackage{caption} 
\usepackage{url}            
\usepackage{booktabs}       
\usepackage{amsfonts}       
\usepackage{nicefrac}       
\usepackage{microtype}      
\usepackage{xcolor}         
\usepackage{lipsum} 
\usepackage{hhline}
\usepackage{boldline}
\usepackage{multirow}
\usepackage{pifont}
\usepackage{enumitem}
\usepackage{caption}
\usepackage{graphicx} 
\usepackage{bm}
\usepackage{algorithm,algorithmic}
\usepackage{amsmath}
\usepackage{colortbl}
\usepackage{graphicx}
\usepackage{subfigure}
\usepackage{footmisc} 
\usepackage{xcolor} 
\usepackage{dsfont}
\usepackage{adjustbox}
\usepackage{booktabs}    
\usepackage{colortbl}
\usepackage{algorithm}
\usepackage{algorithmic}

\usepackage{newfloat}
\usepackage{listings}
\DeclareCaptionStyle{ruled}{labelfont=normalfont,labelsep=colon,strut=off} 
\floatstyle{ruled}
\newfloat{listing}{tb}{lst}{}
\floatname{listing}{Listing}
\title{Adaptive Evidential Learning for Temporal-Semantic Robustness in Moment Retrieval}
\author{
    Haojian Huang\textsuperscript{\rm 1,2}\thanks{Equal Contribution}, 
    Kaijing Ma\textsuperscript{\rm 2,3}\footnotemark[1], 
    Jin Chen\textsuperscript{\rm 2,3,5}\footnotemark[1], 
    Haodong Chen\textsuperscript{\rm 1}, 
    Zhou Wu, 
    Xianghao Zang\textsuperscript{\rm 2}, \\ 
    Han Fang\textsuperscript{\rm 2}, 
    Chao Ban\textsuperscript{\rm 2}, 
    Hao Sun\textsuperscript{\rm 2}\thanks{Corresponding Author},
    Mulin Chen\textsuperscript{\rm 4,2}\footnotemark[2], 
    Zhongjiang He\textsuperscript{\rm 2}
}
\affiliations{
    \textsuperscript{\rm 1}Hong Kong University of Science and Technology (Guangzhou), \quad
    \textsuperscript{\rm 2}Institute of Artificial Intelligence (TeleAl), China Telecom \\
    \textsuperscript{\rm 3}Fudan University, \quad
    \textsuperscript{\rm 4}Northwestern Polytechnical University, \quad
    \textsuperscript{\rm 5}Shanghai Innovation Institute
}

\usepackage{bibentry}

\begin{document}
\input{Contents/math_commands}
\definecolor{myorange}{RGB}{254,236,176} 
\newcommand{\gray}{\rowcolor[gray]{.90}}
\newcommand{\rmnum}[1]{\romannumeral #1}
\newcommand{\Rmnum}[1]{\expandafter\@slowromancap\romannumeral #1@}

\newtheorem{remark}{\noindent \textbf{Remark}}
\newtheorem{defn}{\hspace{-1mm} Definition}

\newcommand{\cmark}{{\textcolor{c1}\Checkmark}}
\newcommand{\xmark}{{\textcolor{c2}\XSolidBrush}}

\newcommand{\hhj}[1]{{\color{orange}#1}}
\newcommand{\todo}[1]{{\color{red}[Todo: {#1}]}}
\newcommand{\RomanNumeralCaps}[1]{\MakeUppercase{\romannumeral #1}}

\newcommand*\diff{\mathop{}\!\mathrm{d}}
\newcommand{\norm}[1]{\left\lVert#1\right\rVert}

\renewcommand{\thefootnote}{\arabic{footnote}}

\newcommand{\beginsupplement}{%
    \setcounter{table}{0}
    \renewcommand{\thetable}{S\arabic{table}}%
    \renewcommand{\theHtable}{Supplement.\thetable}

    \setcounter{algorithm}{0}
    \renewcommand{\thealgorithm}{S\arabic{algorithm}}%
    \renewcommand{\theHalgorithm}{Supplement.\thealgorithm}

    \setcounter{equation}{0}
    \renewcommand{\theequation}{S\arabic{equation}}%
    \renewcommand{\theHequation}{Supplement.\theequation}

    \setcounter{figure}{0}
    \renewcommand{\thefigure}{S\arabic{figure}}%
    \renewcommand{\theHfigure}{Supplement.\thefigure}

    \setcounter{section}{0}
    \renewcommand{\thesection}{S\arabic{section}}%
    \renewcommand{\theHsection}{Supplement.\thesection}
}

\newcommand\blfootnote[1]{%
  \begingroup
  \renewcommand\thefootnote{}\footnote{#1}%
  \addtocounter{footnote}{-1}%
  \endgroup
}

\renewcommand*{\thefootnote}{(\arabic{footnote})}

\newcommand{\qcr}[1]{{\fontfamily{cmtt}\selectfont #1}}
\newcommand{\nig}{\text{\qcr{NIG}}} 
\newcommand{\unit}{clip}
\newcommand{\KLD}[2]{\ensuremath{D_{KL}\infdivx{#1}{#2}}\xspace}

\newcommand{\Loss}{\mathcal{L}}
\newcommand{\Lnll}{\Loss^{\scalebox{.7}{{NLL}}}}
\newcommand{\Lreg}{\Loss^{\scalebox{.7}{{R}}}}
\newcommand{\fix}{\marginpar{FIX}}
\newcommand{\new}{\marginpar{NEW}}
\maketitle

\begin{abstract}
In the domain of moment retrieval, accurately identifying temporal segments within videos based on natural language queries remains challenging. Traditional methods often employ pre-trained models that struggle with fine-grained information and deterministic reasoning, leading to difficulties in aligning with complex or ambiguous moments. To overcome these limitations, we explore Deep Evidential Regression (DER) to construct a vanilla Evidential baseline. However, this approach encounters two major issues: the inability to effectively handle modality imbalance and the structural differences in DER's heuristic uncertainty regularizer, which adversely affect uncertainty estimation. This misalignment results in high uncertainty being incorrectly associated with accurate samples rather than challenging ones. Our observations indicate that existing methods lack the adaptability required for complex video scenarios. In response, we propose  \underline{D}ebiased \underline{E}vidential Learning for \underline{M}oment \underline{R}etrieval (\textbf{\texttt{DEMR}}), a novel framework that incorporates a Reflective Flipped Fusion (RFF) block for cross-modal alignment and a query reconstruction task to enhance text sensitivity, thereby reducing bias in uncertainty estimation. Additionally, we introduce a Geom-regularizer to refine uncertainty predictions, enabling adaptive alignment with difficult moments and improving retrieval accuracy. Extensive testing on standard datasets and debiased datasets ActivityNet-CD and Charades-CD demonstrates significant enhancements in effectiveness, robustness, and interpretability, positioning our approach as a promising solution for temporal-semantic robustness in moment retrieval. 
The code is publicly available at \textcolor{red}{https://github.com/KaijingOfficial/DEMR}.
\end{abstract}


\begin{figure}[h]
    \centering
    \includegraphics[width=0.47\textwidth]{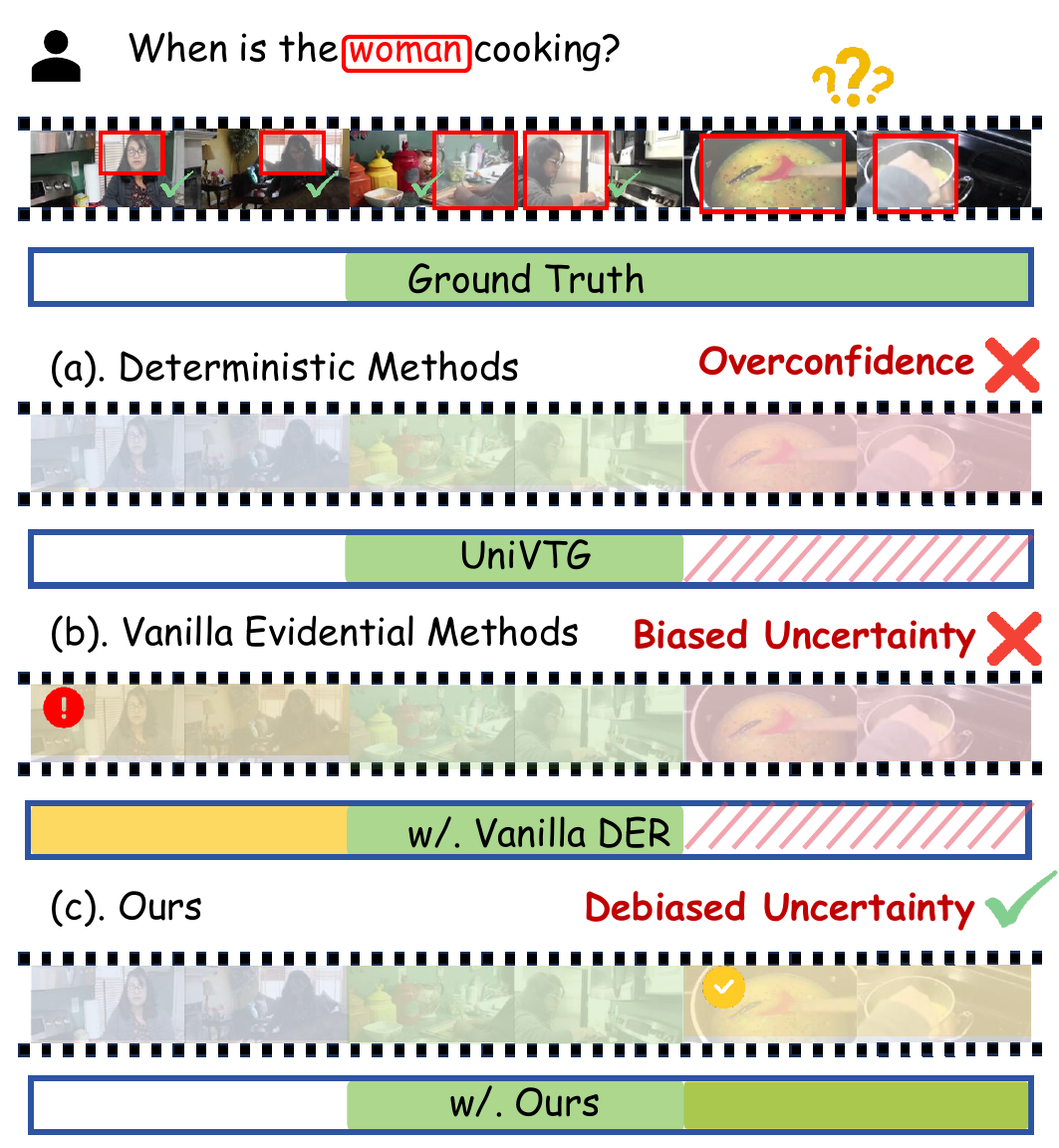}
    \vspace{-2em}
    \caption{Comparison of MR methods: (a) Deterministic methods (\emph{e.g.}~\cite{lin2023univtg}) are overconfident with limited evidence, using Non-Maximum Suppression (NMS) but still failing on challenging frames; (b) Vanilla evidential methods consider uncertainty but produce biased estimates on hard samples; (c) Our method adaptively aligns with challenging semantics for accurate uncertainty modeling and improved inference. \textbf{\textcolor{myorange}{Yellow}} regions denotes uncertainty predictions.}
    \label{fig:comp}
    \vspace{-2em}
\end{figure}
\input{Contents/Intro}

\input{Contents/Related_Work}

\input{Contents/Preliminaries}

\input{Contents/Methdology}

\input{Contents/Exp_new}

\input{Contents/Conclusion}

\bibliography{aaai2026}


\end{document}

%% file: Contents/math_commands.tex


\newcommand{\figleft}{{\em (Left)}}
\newcommand{\figcenter}{{\em (Center)}}
\newcommand{\figright}{{\em (Right)}}
\newcommand{\figtop}{{\em (Top)}}
\newcommand{\figbottom}{{\em (Bottom)}}
\newcommand{\captiona}{{\em (a)}}
\newcommand{\captionb}{{\em (b)}}
\newcommand{\captionc}{{\em (c)}}
\newcommand{\captiond}{{\em (d)}}

\newcommand{\newterm}[1]{{\bf #1}}

\def\figref#1{figure~\ref{#1}}
\def\Figref#1{Figure~\ref{#1}}
\def\twofigref#1#2{figures \ref{#1} and \ref{#2}}
\def\quadfigref#1#2#3#4{figures \ref{#1}, \ref{#2}, \ref{#3} and \ref{#4}}
\def\secref#1{section~\ref{#1}}
\def\Secref#1{Section~\ref{#1}}
\def\twosecrefs#1#2{sections \ref{#1} and \ref{#2}}
\def\secrefs#1#2#3{sections \ref{#1}, \ref{#2} and \ref{#3}}
\def\eqref#1{equation~\ref{#1}}
\def\Eqref#1{Equation~\ref{#1}}
\def\plaineqref#1{\ref{#1}}
\def\chapref#1{chapter~\ref{#1}}
\def\Chapref#1{Chapter~\ref{#1}}
\def\rangechapref#1#2{chapters\ref{#1}--\ref{#2}}
\def\algref#1{algorithm~\ref{#1}}
\def\Algref#1{Algorithm~\ref{#1}}
\def\twoalgref#1#2{algorithms \ref{#1} and \ref{#2}}
\def\Twoalgref#1#2{Algorithms \ref{#1} and \ref{#2}}
\def\partref#1{part~\ref{#1}}
\def\Partref#1{Part~\ref{#1}}
\def\twopartref#1#2{parts \ref{#1} and \ref{#2}}

\def\ceil#1{\lceil #1 \rceil}
\def\floor#1{\lfloor #1 \rfloor}
\def\1{\bm{1}}
\newcommand{\train}{\mathcal{D}}
\newcommand{\valid}{\mathcal{D_{\mathrm{valid}}}}
\newcommand{\test}{\mathcal{D_{\mathrm{test}}}}

\def\eps{{\epsilon}}

\def\reta{{\textnormal{$\eta$}}}
\def\ra{{\textnormal{a}}}
\def\rb{{\textnormal{b}}}
\def\rc{{\textnormal{c}}}
\def\rd{{\textnormal{d}}}
\def\re{{\textnormal{e}}}
\def\rf{{\textnormal{f}}}
\def\rg{{\textnormal{g}}}
\def\rh{{\textnormal{h}}}
\def\ri{{\textnormal{i}}}
\def\rj{{\textnormal{j}}}
\def\rk{{\textnormal{k}}}
\def\rl{{\textnormal{l}}}
\def\rn{{\textnormal{n}}}
\def\ro{{\textnormal{o}}}
\def\rp{{\textnormal{p}}}
\def\rq{{\textnormal{q}}}
\def\rr{{\textnormal{r}}}
\def\rs{{\textnormal{s}}}
\def\rt{{\textnormal{t}}}
\def\ru{{\textnormal{u}}}
\def\rv{{\textnormal{v}}}
\def\rw{{\textnormal{w}}}
\def\rx{{\textnormal{x}}}
\def\ry{{\textnormal{y}}}
\def\rz{{\textnormal{z}}}

\def\rvepsilon{{\mathbf{\epsilon}}}
\def\rvtheta{{\mathbf{\theta}}}
\def\rva{{\mathbf{a}}}
\def\rvb{{\mathbf{b}}}
\def\rvc{{\mathbf{c}}}
\def\rvd{{\mathbf{d}}}
\def\rve{{\mathbf{e}}}
\def\rvf{{\mathbf{f}}}
\def\rvg{{\mathbf{g}}}
\def\rvh{{\mathbf{h}}}
\def\rvu{{\mathbf{i}}}
\def\rvj{{\mathbf{j}}}
\def\rvk{{\mathbf{k}}}
\def\rvl{{\mathbf{l}}}
\def\rvm{{\mathbf{m}}}
\def\rvn{{\mathbf{n}}}
\def\rvo{{\mathbf{o}}}
\def\rvp{{\mathbf{p}}}
\def\rvq{{\mathbf{q}}}
\def\rvr{{\mathbf{r}}}
\def\rvs{{\mathbf{s}}}
\def\rvt{{\mathbf{t}}}
\def\rvu{{\mathbf{u}}}
\def\rvv{{\mathbf{v}}}
\def\rvw{{\mathbf{w}}}
\def\rvx{{\mathbf{x}}}
\def\rvy{{\mathbf{y}}}
\def\rvz{{\mathbf{z}}}

\def\erva{{\textnormal{a}}}
\def\ervb{{\textnormal{b}}}
\def\ervc{{\textnormal{c}}}
\def\ervd{{\textnormal{d}}}
\def\erve{{\textnormal{e}}}
\def\ervf{{\textnormal{f}}}
\def\ervg{{\textnormal{g}}}
\def\ervh{{\textnormal{h}}}
\def\ervi{{\textnormal{i}}}
\def\ervj{{\textnormal{j}}}
\def\ervk{{\textnormal{k}}}
\def\ervl{{\textnormal{l}}}
\def\ervm{{\textnormal{m}}}
\def\ervn{{\textnormal{n}}}
\def\ervo{{\textnormal{o}}}
\def\ervp{{\textnormal{p}}}
\def\ervq{{\textnormal{q}}}
\def\ervr{{\textnormal{r}}}
\def\ervs{{\textnormal{s}}}
\def\ervt{{\textnormal{t}}}
\def\ervu{{\textnormal{u}}}
\def\ervv{{\textnormal{v}}}
\def\ervw{{\textnormal{w}}}
\def\ervx{{\textnormal{x}}}
\def\ervy{{\textnormal{y}}}
\def\ervz{{\textnormal{z}}}

\def\rmA{{\mathbf{A}}}
\def\rmB{{\mathbf{B}}}
\def\rmC{{\mathbf{C}}}
\def\rmD{{\mathbf{D}}}
\def\rmE{{\mathbf{E}}}
\def\rmF{{\mathbf{F}}}
\def\rmG{{\mathbf{G}}}
\def\rmH{{\mathbf{H}}}
\def\rmI{{\mathbf{I}}}
\def\rmJ{{\mathbf{J}}}
\def\rmK{{\mathbf{K}}}
\def\rmL{{\mathbf{L}}}
\def\rmM{{\mathbf{M}}}
\def\rmN{{\mathbf{N}}}
\def\rmO{{\mathbf{O}}}
\def\rmP{{\mathbf{P}}}
\def\rmQ{{\mathbf{Q}}}
\def\rmR{{\mathbf{R}}}
\def\rmS{{\mathbf{S}}}
\def\rmT{{\mathbf{T}}}
\def\rmU{{\mathbf{U}}}
\def\rmV{{\mathbf{V}}}
\def\rmW{{\mathbf{W}}}
\def\rmX{{\mathbf{X}}}
\def\rmY{{\mathbf{Y}}}
\def\rmZ{{\mathbf{Z}}}

\def\ermA{{\textnormal{A}}}
\def\ermB{{\textnormal{B}}}
\def\ermC{{\textnormal{C}}}
\def\ermD{{\textnormal{D}}}
\def\ermE{{\textnormal{E}}}
\def\ermF{{\textnormal{F}}}
\def\ermG{{\textnormal{G}}}
\def\ermH{{\textnormal{H}}}
\def\ermI{{\textnormal{I}}}
\def\ermJ{{\textnormal{J}}}
\def\ermK{{\textnormal{K}}}
\def\ermL{{\textnormal{L}}}
\def\ermM{{\textnormal{M}}}
\def\ermN{{\textnormal{N}}}
\def\ermO{{\textnormal{O}}}
\def\ermP{{\textnormal{P}}}
\def\ermQ{{\textnormal{Q}}}
\def\ermR{{\textnormal{R}}}
\def\ermS{{\textnormal{S}}}
\def\ermT{{\textnormal{T}}}
\def\ermU{{\textnormal{U}}}
\def\ermV{{\textnormal{V}}}
\def\ermW{{\textnormal{W}}}
\def\ermX{{\textnormal{X}}}
\def\ermY{{\textnormal{Y}}}
\def\ermZ{{\textnormal{Z}}}

\def\vzero{{\bm{0}}}
\def\vone{{\bm{1}}}
\def\vmu{{\bm{\mu}}}
\def\vtheta{{\bm{\theta}}}
\def\va{{\bm{a}}}
\def\vb{{\bm{b}}}
\def\vc{{\bm{c}}}
\def\vd{{\bm{d}}}
\def\ve{{\bm{e}}}
\def\vf{{\bm{f}}}
\def\vg{{\bm{g}}}
\def\vh{{\bm{h}}}
\def\vi{{\bm{i}}}
\def\vj{{\bm{j}}}
\def\vk{{\bm{k}}}
\def\vl{{\bm{l}}}
\def\vm{{\bm{m}}}
\def\vn{{\bm{n}}}
\def\vo{{\bm{o}}}
\def\vp{{\bm{p}}}
\def\vq{{\bm{q}}}
\def\vr{{\bm{r}}}
\def\vs{{\bm{s}}}
\def\vt{{\bm{t}}}
\def\vu{{\bm{u}}}
\def\vv{{\bm{v}}}
\def\vw{{\bm{w}}}
\def\vx{{\bm{x}}}
\def\vy{{\bm{y}}}
\def\vz{{\bm{z}}}

\def\evalpha{{\alpha}}
\def\evbeta{{\beta}}
\def\evepsilon{{\epsilon}}
\def\evlambda{{\lambda}}
\def\evomega{{\omega}}
\def\evmu{{\mu}}
\def\evpsi{{\psi}}
\def\evsigma{{\sigma}}
\def\evtheta{{\theta}}
\def\eva{{a}}
\def\evb{{b}}
\def\evc{{c}}
\def\evd{{d}}
\def\eve{{e}}
\def\evf{{f}}
\def\evg{{g}}
\def\evh{{h}}
\def\evi{{i}}
\def\evj{{j}}
\def\evk{{k}}
\def\evl{{l}}
\def\evm{{m}}
\def\evn{{n}}
\def\evo{{o}}
\def\evp{{p}}
\def\evq{{q}}
\def\evr{{r}}
\def\evs{{s}}
\def\evt{{t}}
\def\evu{{u}}
\def\evv{{v}}
\def\evw{{w}}
\def\evx{{x}}
\def\evy{{y}}
\def\evz{{z}}

\def\mA{{\bm{A}}}
\def\mB{{\bm{B}}}
\def\mC{{\bm{C}}}
\def\mD{{\bm{D}}}
\def\mE{{\bm{E}}}
\def\mF{{\bm{F}}}
\def\mG{{\bm{G}}}
\def\mH{{\bm{H}}}
\def\mI{{\bm{I}}}
\def\mJ{{\bm{J}}}
\def\mK{{\bm{K}}}
\def\mL{{\bm{L}}}
\def\mM{{\bm{M}}}
\def\mN{{\bm{N}}}
\def\mO{{\bm{O}}}
\def\mP{{\bm{P}}}
\def\mQ{{\bm{Q}}}
\def\mR{{\bm{R}}}
\def\mS{{\bm{S}}}
\def\mT{{\bm{T}}}
\def\mU{{\bm{U}}}
\def\mV{{\bm{V}}}
\def\mW{{\bm{W}}}
\def\mX{{\bm{X}}}
\def\mY{{\bm{Y}}}
\def\mZ{{\bm{Z}}}
\def\mBeta{{\bm{\beta}}}
\def\mPhi{{\bm{\Phi}}}
\def\mLambda{{\bm{\Lambda}}}
\def\mSigma{{\bm{\Sigma}}}

\newcommand{\tens}[1]{\bm{\mathsfit{#1}}}
\def\tA{{\tens{A}}}
\def\tB{{\tens{B}}}
\def\tC{{\tens{C}}}
\def\tD{{\tens{D}}}
\def\tE{{\tens{E}}}
\def\tF{{\tens{F}}}
\def\tG{{\tens{G}}}
\def\tH{{\tens{H}}}
\def\tI{{\tens{I}}}
\def\tJ{{\tens{J}}}
\def\tK{{\tens{K}}}
\def\tL{{\tens{L}}}
\def\tM{{\tens{M}}}
\def\tN{{\tens{N}}}
\def\tO{{\tens{O}}}
\def\tP{{\tens{P}}}
\def\tQ{{\tens{Q}}}
\def\tR{{\tens{R}}}
\def\tS{{\tens{S}}}
\def\tT{{\tens{T}}}
\def\tU{{\tens{U}}}
\def\tV{{\tens{V}}}
\def\tW{{\tens{W}}}
\def\tX{{\tens{X}}}
\def\tY{{\tens{Y}}}
\def\tZ{{\tens{Z}}}

\def\gA{{\mathcal{A}}}
\def\gB{{\mathcal{B}}}
\def\gC{{\mathcal{C}}}
\def\gD{{\mathcal{D}}}
\def\gE{{\mathcal{E}}}
\def\gF{{\mathcal{F}}}
\def\gG{{\mathcal{G}}}
\def\gH{{\mathcal{H}}}
\def\gI{{\mathcal{I}}}
\def\gJ{{\mathcal{J}}}
\def\gK{{\mathcal{K}}}
\def\gL{{\mathcal{L}}}
\def\gM{{\mathcal{M}}}
\def\gN{{\mathcal{N}}}
\def\gO{{\mathcal{O}}}
\def\gP{{\mathcal{P}}}
\def\gQ{{\mathcal{Q}}}
\def\gR{{\mathcal{R}}}
\def\gS{{\mathcal{S}}}
\def\gT{{\mathcal{T}}}
\def\gU{{\mathcal{U}}}
\def\gV{{\mathcal{V}}}
\def\gW{{\mathcal{W}}}
\def\gX{{\mathcal{X}}}
\def\gY{{\mathcal{Y}}}
\def\gZ{{\mathcal{Z}}}

\def\sA{{\mathbb{A}}}
\def\sB{{\mathbb{B}}}
\def\sC{{\mathbb{C}}}
\def\sD{{\mathbb{D}}}
\def\sF{{\mathbb{F}}}
\def\sG{{\mathbb{G}}}
\def\sH{{\mathbb{H}}}
\def\sI{{\mathbb{I}}}
\def\sJ{{\mathbb{J}}}
\def\sK{{\mathbb{K}}}
\def\sL{{\mathbb{L}}}
\def\sM{{\mathbb{M}}}
\def\sN{{\mathbb{N}}}
\def\sO{{\mathbb{O}}}
\def\sP{{\mathbb{P}}}
\def\sQ{{\mathbb{Q}}}
\def\sR{{\mathbb{R}}}
\def\sS{{\mathbb{S}}}
\def\sT{{\mathbb{T}}}
\def\sU{{\mathbb{U}}}
\def\sV{{\mathbb{V}}}
\def\sW{{\mathbb{W}}}
\def\sX{{\mathbb{X}}}
\def\sY{{\mathbb{Y}}}
\def\sZ{{\mathbb{Z}}}

\def\emLambda{{\Lambda}}
\def\emA{{A}}
\def\emB{{B}}
\def\emC{{C}}
\def\emD{{D}}
\def\emE{{E}}
\def\emF{{F}}
\def\emG{{G}}
\def\emH{{H}}
\def\emI{{I}}
\def\emJ{{J}}
\def\emK{{K}}
\def\emL{{L}}
\def\emM{{M}}
\def\emN{{N}}
\def\emO{{O}}
\def\emP{{P}}
\def\emQ{{Q}}
\def\emR{{R}}
\def\emS{{S}}
\def\emT{{T}}
\def\emU{{U}}
\def\emV{{V}}
\def\emW{{W}}
\def\emX{{X}}
\def\emY{{Y}}
\def\emZ{{Z}}
\def\emSigma{{\Sigma}}

\newcommand{\etens}[1]{\mathsfit{#1}}
\def\etLambda{{\etens{\Lambda}}}
\def\etA{{\etens{A}}}
\def\etB{{\etens{B}}}
\def\etC{{\etens{C}}}
\def\etD{{\etens{D}}}
\def\etE{{\etens{E}}}
\def\etF{{\etens{F}}}
\def\etG{{\etens{G}}}
\def\etH{{\etens{H}}}
\def\etI{{\etens{I}}}
\def\etJ{{\etens{J}}}
\def\etK{{\etens{K}}}
\def\etL{{\etens{L}}}
\def\etM{{\etens{M}}}
\def\etN{{\etens{N}}}
\def\etO{{\etens{O}}}
\def\etP{{\etens{P}}}
\def\etQ{{\etens{Q}}}
\def\etR{{\etens{R}}}
\def\etS{{\etens{S}}}
\def\etT{{\etens{T}}}
\def\etU{{\etens{U}}}
\def\etV{{\etens{V}}}
\def\etW{{\etens{W}}}
\def\etX{{\etens{X}}}
\def\etY{{\etens{Y}}}
\def\etZ{{\etens{Z}}}

\newcommand{\pdata}{p_{\rm{data}}}
\newcommand{\ptrain}{\hat{p}_{\rm{data}}}
\newcommand{\Ptrain}{\hat{P}_{\rm{data}}}
\newcommand{\pmodel}{p_{\rm{model}}}
\newcommand{\Pmodel}{P_{\rm{model}}}
\newcommand{\ptildemodel}{\tilde{p}_{\rm{model}}}
\newcommand{\pencode}{p_{\rm{encoder}}}
\newcommand{\pdecode}{p_{\rm{decoder}}}
\newcommand{\precons}{p_{\rm{reconstruct}}}

\newcommand{\E}{\mathbb{E}}
\newcommand{\Ls}{\mathcal{L}}
\newcommand{\R}{\mathbb{R}}
\newcommand{\emp}{\tilde{p}}
\newcommand{\lr}{\alpha}
\newcommand{\reg}{\lambda}
\newcommand{\rect}{\mathrm{rectifier}}
\newcommand{\softmax}{\mathrm{softmax}}
\newcommand{\sigmoid}{\sigma}
\newcommand{\softplus}{\zeta}
\newcommand{\KL}{D_{\mathrm{KL}}}
\newcommand{\Var}{\mathrm{Var}}
\newcommand{\standarderror}{\mathrm{SE}}
\newcommand{\Cov}{\mathrm{Cov}}
\newcommand{\normlzero}{L^0}
\newcommand{\normlone}{L^1}
\newcommand{\normltwo}{L^2}
\newcommand{\normlp}{L^p}
\newcommand{\normmax}{L^\infty}

\newcommand{\parents}{Pa} 

\let\ab\allowbreak

%% file: Contents/Intro.tex
\section{Introduction}
\label{sec:introduction}
Moment Retrieval (MR) in video understanding is a pivotal task that involves locating specific time segments within untrimmed videos based on natural language queries~\citep{hu2024recent}. This task demands models to effectively integrate and interpret both visual and textual information to accurately identify relevant moments. Traditional approaches~\cite{lin2023univtg,liu2024r2-tuning,lei2021detecting_Moment-DETR,li2024momentdiff,moon2023query_QD-DETR,huang2022video,chen2024uncertainty,huang2025vistadpo,zhang2025learning,liu2025sample,chen2023transformer} predominantly rely on pretrained Transformer models such as CLIP-ViT~\cite{Radford2021LearningTV}, which are primarily pretrained on instance-level classification, thereby limiting their ability to address the nuanced demands of MR in complex video scenarios. Even when models enhance their ability to align visual and textual features, they often operate under deterministic paradigms. These methods struggle to provide accurate predictions in challenging video frames. As illustrated in Figure~\ref{fig:comp}, when frames lack the presence of a woman, models find it difficult to align with the query "cooking." And during inference, techniques like Non-Maximum Suppression (NMS) are used to select the most probable segments. But without additional knowledge, these methods fail to align with hard moments. To address this, we integrate Deep Evidential Regression (DER)~\cite{amini2020deep} into a vanilla evidential method as a baseline (\emph{i.e.} Figure~\ref{fig:comp}(b)). Based on Evidential Deep Learning (EDL)~\cite{sensoy2018evidential}, DER excels at capturing uncertainty by learning a second-order distribution to fit the correct moment proposals. Unlike deterministic methods, DER represents uncertainty by treating each proposal as evidence sampled from observations. This approach learns the correspondence between evidence and ground truth, aligning a second-order distribution to the target domain. Through this learned distribution, the model assesses sample uncertainty and adjusts gradient backpropagation, resulting in more robust and accurate inference. This capability has already led to significant success in various regression tasks~\cite{wang2022uncertainty,wu2023estimating,ye2024uncertainty,ma2024beyond}.

However, applying DER directly to MR presents challenges. Unlike other unimodal domains~\cite{wu2023estimating,ye2024uncertainty}, evidence requires comprehensive fusion of visual and textual information in MR.
Simple concatenation of multimodal features fails to ensure nuanced understanding. Additionally, DER faces counterintuitive uncertainty issues: higher-error predictions can receive lower uncertainty due to limitations in vanilla DER's regularizer. Unlike classification-based EDL methods, DER lacks a standard KL-divergence term, relying on a heuristic regularizer that overly suppresses evidence, especially in low-error samples, misaligning uncertainty estimates with low-error samples showing higher uncertainty and vice versa.

To address modality imbalance, we propose \underline{D}ebiased \underline{E}vidential Learning for \underline{M}oment \underline{R}etrieval (\textbf{\texttt{DEMR}}). Our approach incorporates a Reflective Flipped Fusion (RFF) block with dual branches for progressive cross-modal alignment, complemented by a query reconstruction (QR) task to strengthen the text branch. This design enhances the model's sensitivity to textual information and mitigates bias in uncertainty estimation. To resolve counterintuitive uncertainty, we introduce a simple yet effective Geom-regularizer, which adjusts uncertainty estimation based on prediction accuracy, adaptively suppressing overconfidence and debiasing the system. Our contributions are summarized as follows:
\vspace{-1.5mm}
\begin{itemize}[leftmargin=*]
    \item We introduce an uncertainty-aware MR baseline with DER, and further develop DEMR to address challenging and ambiguous moments.
    \item Our approach integrates the RFF block, auxiliary QR task, and Geom-regularizer to mitigate modality imbalance and improve uncertainty estimation, enabling adaptive alignment with difficult moments.
    \item We conducted experiments on common moment retrieval datasets and debiased versions of ActivityNet-CD and Charades-STA CD. Results show our method's effectiveness, robustness, and interpretability across benchmarks.
\end{itemize}

%% file: Contents/Related_Work.tex
\section{Related Work}
\label{sec:related_work}

Moment Retrieval focuses on localizing temporal segments in untrimmed videos based on textual queries, and has evolved rapidly in recent years. Early MR methods adopted two-stage frameworks~\cite{zhang2019exploiting_TCMN, zhang2019cross_QSPN, gao2021fast_SAP}, generating temporal proposals followed by refinement steps such as Non-Maximum Suppression (NMS), but these approaches were computationally expensive and relied heavily on hand-crafted priors. One-stage models~\cite{chen2018temporally_tgn, wang2020temporally_cbp, otani2020uncovering_SCDM, Zhang_2019_CVPR_man, hu2021video_CMHN, liu2018temporal_tmn, zhang2020learning_2d_tan} aimed to directly predict moment boundaries, improving efficiency but still faced challenges in flexibility and alignment. The introduction of transformer-based architectures, notably Detection Transformer (DETR)~\cite{carion2020end_DETR}, reframed MR as a set prediction problem, inspiring models such as Moment-DETR~\cite{lei2021detecting_Moment-DETR}, QD-DETR~\cite{moon2023query_QD-DETR}, and MESM~\cite{liu2024towards_MESM}, which further advanced cross-modal alignment and prediction accuracy. However, despite these advances, most state-of-the-art MR methods remain deterministic and lack effective mechanisms for modeling uncertainty, limiting their robustness in complex or ambiguous scenarios.

Recently, uncertainty Learning has gained attention as MR datasets are found to contain inherent ambiguities and biases~\cite{zhang2023temporal}. Annotation uncertainty arises from inconsistent temporal boundaries set by different annotators, while query uncertainty stems from diverse natural language descriptions for the same video moment. Additionally, dataset bias—such as overrepresentation of common events and long-tailed distributions—further challenges model generalization~\cite{zhang2023temporal, otani2020uncovering}. These issues highlight the need for uncertainty-aware modeling. Evidential Deep Learning (EDL), grounded in Dempster-Shafer Theory~\cite{shafer1992dempster} and Subjective Logic~\cite{sensoy2018evidential, josang2016subjective}, explicitly models uncertainty via second-order probability distributions and has shown effectiveness in classification~\cite{bao2021evidential, han2020trusted, han2022trusted, chen2024towards, huang2024evidential, huang2024trusted, chen2024bovila, holmquist2023evidential, liu2024adaptive,huang2024crest,huang2025structure} and regression tasks~\cite{amini2020deep, wang2022uncertainty, wu2023estimating}. Deep Evidential Regression (DER) extends EDL to regression, but suffers from evidence contraction and gradient issues under high uncertainty~\cite{wu2024evidence, ye2024uncertainty}. Recent advances have introduced new regularizers to address these limitations, but their application to MR remains unexplored. In this work, we bridge this gap by introducing DER into the MR task, aiming to improve model robustness and reliability under uncertainty. We further address modality imbalance and structural limitations of vanilla DER, marking the first successful extension of evidential regression to moment retrieval.

%% file: Contents/Preliminaries.tex
\section{Preliminaries}

DER~\cite{amini2020deep} places evidential priors over the original Gaussian likelihood function and trains the model to infer the hyperparameters of the evidential distribution. This approach enables the model to learn both aleatoric and epistemic uncertainty. 
In our context, adjacent video frames often exhibit similar semantics, which introduces uncertainty in precisely locating temporal boundaries. The start or end temporal boundary of a video is represented by distinct Gaussian distributions: $\bm b \sim\mathcal{N}(\mu,\sigma^2)$, where $\bm b \in \mathbb{R}^{1 \times \mathcal{H}}$ represents the start or end of moments observed $\mathcal{H}$ times. We assume that observations of the same type (either all starts or all ends) are \emph{i.i.d.}. 
 The corresponding expectation $\mu$ and variance $\sigma^2$ of the Gaussian distribution subject to \nig~prior:
\begin{align}
p(\mu,\sigma^2\mid\underbrace{\gamma,\upsilon,\alpha,\beta}_{\boldsymbol{\varphi }})
&=
\mathcal{N}(\mu|\gamma,\sigma^2 \upsilon^{-1})
\Gamma^{-1}(\sigma^2|\alpha,\beta),
\end{align}
where $\boldsymbol{\varphi}=(\gamma, \upsilon, \alpha, \beta)$ are the prior \nig~distribution parameters derived from the video content and user queries, serve as conditionals for the Gaussian estimates of \( b_i \) with $\gamma \in \mathbb{R}, \upsilon > 0, \alpha > 1, \beta > 0$. The gamma function is denoted by $\Gamma(\cdot)$. We use a linear evidential predictor to estimate $\boldsymbol{\varphi }$, training it to maximize the likelihood. 
Since the likelihood function has a form of Student-t distribution ($\mathrm{St}$), we minimize the negative logarithmic likelihood (NLL) as follows: 
\begin{equation}
\label{NLL_Loss}
\mathcal{L}^{\mathrm{NLL}}_{i}=
-\log p(b_i|\boldsymbol{\varphi })=
-\log\left(\mathrm{St}\left(b_i;\gamma,\frac{\beta(1+\upsilon)}{\upsilon\alpha},2\alpha\right)\right),
\end{equation}
Models optimized only on observed samples with the NLL loss (\emph{i.e.} Eq.~\ref{NLL_Loss}) tend to overfit and exhibit overconfidence. To counter this, DER introduced a regularizer for the $i$-th prediction as follows:
\begin{equation}
\label{vallina_reg}
\mathcal{L}^\mathrm{R}_i(\boldsymbol{\vartheta})=\Delta\cdot\Phi,
\end{equation}
where $\Delta = |b_i-\gamma|$ represents the error, $\Phi = 2\upsilon+\alpha$ denotes the evidence, and $\boldsymbol{\vartheta}$ are the model parameters, with $b_i$ as the ground truth. 
Detailed formulation can be found in \textbf{Supplementary Material}.
Using the \nig~distribution, prediction, aleatoric and epistemic uncertainties are calculated as follows:
\begin{equation}
\underbrace{\mathbb{E}[\mu] = \gamma}_{\text{prediction}},\quad\underbrace{\mathbb{E}[\sigma^2]=\frac{\beta}{\alpha-1}}_{\text{aleatoric}}, \quad\underbrace{\mathrm{Var}[\mu]=\frac{\beta}{\upsilon(\alpha-1)}}_{\text{epistemic}},
\end{equation}
$\mathbb{E}[\sigma^2]$ refers to the inherent noise in the data, which cannot be reduced or eliminated. $\mathrm{Var}[\mu]$ reflects the model's lack of confidence in its own predictions due to limited knowledge.

%% file: Contents/Methdology.tex
\begin{figure*}
 \centering
\includegraphics[width=0.96\linewidth]{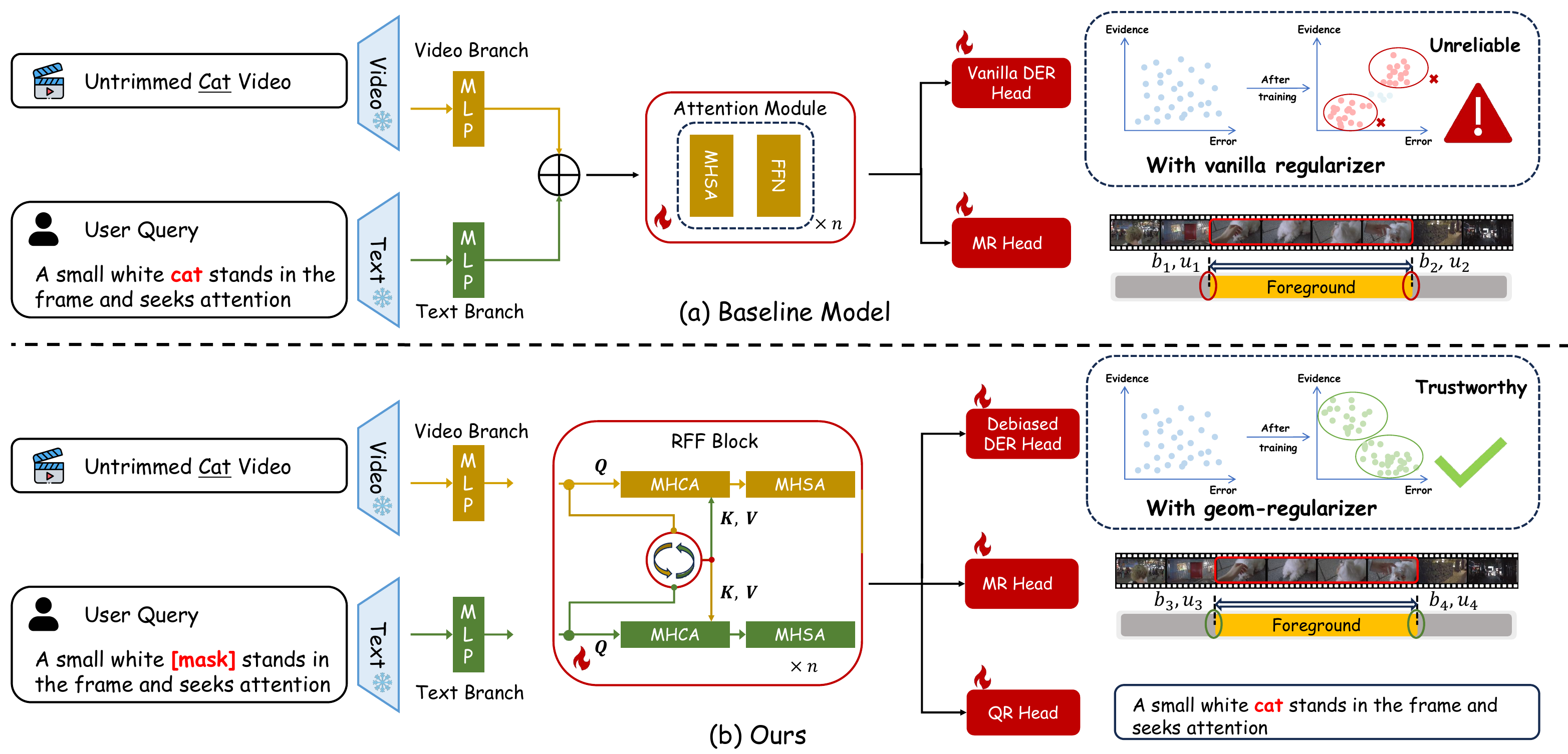}
\vspace{-1em}
\caption{Comparison of the baseline (a) and our improved model (b) for the MR task. In (a), the baseline exhibits weak sensitivity to text, as the overlap between the MR task and DER objective causes over-reliance on visual features, while the vanilla DER regularizer leads to unreliable uncertainty estimates. In (b), our RFF block and QR head enhance cross-modal interaction and text sensitivity, and the Geom-regularizer corrects structural flaws in DER for more reliable uncertainty estimation.}

\label{overview}  
\vspace{-1.8em}
\end{figure*}
\section{Methodology}
\label{sec:method}
\subsection{Problem Definition}
\label{problem}
Given a video \( V = \{v_i\}_{i=1}^{L_v} \) and a query \( Q = \{q_i\}_{i=1}^{L_q} \), both as vectors in \( \mathbb{R}^{D} \), Moment Retrieval aims to locate the time span \( m = [m^s, m^e] \) in the video that best matches \( Q \), by identifying the start and end clips \( m^s \) and \( m^e \) of the relevant segment.


\subsection{Building Baseline with Vanilla DER for MR}
\label{sec:baseline}
Deterministic methods in moment retrieval struggle with nuanced video frames and often fail to align with challenging moments due to their reliance on instance-level classification. In contrast, DER captures uncertainty by learning a second-order distribution, treating each proposal as evidence. This approach enhances robustness and accuracy, allowing for better alignment with complex video scenarios. Therefore, as illustrated in Figure~\ref{overview} (a), we first build an uncertainty-aware baseline by integrating DER into the MR task. The motivation for this is to tackle the challenging video frames in moment retrieval, enabling the model to progressively align difficult samples semantically through uncertainty representation. Overall, the loss function of the model can be formulated as follows:
\begin{equation}
\label{loss:der}
\mathcal{L}^\text{B}_i(\boldsymbol{w}) = 
\lambda_\text{NLL}  \mathcal{L}^{\mathrm{NLL}}_{i} +  \lambda_\text{Reg} \mathcal{L}^\text{R}_i(\boldsymbol{w}), 
\end{equation}
\begin{equation}
\label{loss:all}
\mathcal{L}_{base} =\mathcal{L}_\text{mr} + \lambda_\text{der} \frac{1}{N} \sum_{i=1}^N \mathcal{L}^\text{B}_i(\boldsymbol{w}) ,
\end{equation}
where $N$ symbolizes the number of clips in a training set and $\mathcal{L}_\text{mr}$ denotes MR loss (\emph{i.e.} Eq.~\ref{eq:mr_loss}).
While DER effectively estimates uncertainty, its vanilla form presents limitations like modality imbalance and flawed uncertainty estimation, which our baseline exposes and sets the foundation for improvement.

\subsection{Debiased DER Model for MR}
To address the biased uncertainty estimation in the baseline in section~\ref{sec:baseline}, caused by modality imbalance and counterintuitive uncertainty, we propose \textbf{\texttt{DEMR}}. Our model introduces a RFF block for progressive cross-modal alignment, reducing over-reliance on visual features, and a QR task to enhance text sensitivity. As shown in Figure~\ref{overview} (b), DEMR first encodes an untrimmed video and masked query, reconstructs the masked tokens via the RFF block, and performs MR. The debiased DER head assesses both aleatoric and epistemic uncertainties, while the MR and QR heads manage task stages. Further details are provided in subsequent sections.

\vspace{-0.4em}
\subsubsection*{\textbf{RFF Block.}}
The RFF block iteratively updates video and text features by alternating their roles as queries and keys/values in a shared cross-attention (CA) module, followed by self-attention (SA) refinement in each branch. At each layer, features are updated as:
\begin{equation}
V^{(i+1)} = SA_v^{(i)}(CA_{q \to v}^{(i)}), \quad Q^{(i+1)} = SA_q^{(i)}(CA_{v \to q}^{(i)})
\end{equation}
where CA and SA are defined as:
\begin{align}
CA_{v \to q}^{(i)} &= \text{Softmax}\left(\frac{V^{(i)} Q^{(i)\mathsf{T}}}{\sqrt{d_k}}\right) Q^{(i)} \\
SA_v^{(i)} &= \text{Softmax}\left(\frac{CA_{v \to q}^{(i)} CA_{v \to q}^{(i)\mathsf{T}}}{\sqrt{d_k}}\right) CA_{v \to q}^{(i)}
\end{align}
This process repeats for $n$ layers, progressively enhancing cross-modal alignment.

\vspace{-0.4em}
\subsubsection*{\textbf{MR Head.}} Given $\tilde{\mathbf{V}}^k \in \mathbb{R}^{L_v\times D}$, this head generates a series of offsets $\{\tilde{m}_i\}_{i=1}^{L_v}$ for each unit. We then define the predicted boundary $\tilde{m}_i$ and the corresponding interval $d_i$ (i.e., $d_i = m_i^s - m_i^e$). For training objectives, we use a combination of smooth L1 loss and generalized IoU loss to optimize the model's performance.
\begin{equation}
\label{eq:mr_loss}
    \mathcal{L}_\text{mr} = \mathds1 _{f_i=1}  \left[
    \lambda_\text{L1}\mathcal{L}_{\text{SmoothL1}}\left(\tilde{d}_i, {d_i}\right)+
    \lambda_\text{iou} \mathcal{L}_\text{iou}\left( \tilde{m}_i, {m_i} \right) \right].
\end{equation}

Notably, this regression objective is only devised for foreground \unit s~\textit{i.e.,} $f_i=1$. 
\vspace{-0.4em}
\subsubsection*{\textbf{Query Reconstruction Task.}}
We observe that nouns are crucial for cross-modal tasks, as CLIP features primarily capture objects due to training on static images, leading to a focus on static visual cues rather than dynamic actions. In our benchmarks, identifying key nouns is often sufficient for high-quality reasoning, motivating our QR task design. To enhance cross-modal alignment, we mask query entities at a fixed ratio during early alignment, forcing the model to employ both video context and remaining query tokens. The QR head then reconstructs the masked tokens, with a dedicated loss optimizing cross-modal inference as follows:
\begin{equation}
\mathcal{L}_{qr} = \mathbb{E} \left[ -\sum_{i=1}^{l} \log P(w_i \mid U, V) \right],
\end{equation}
Here, \( l \) is the number of masked tokens, \( w_i \) denotes the \( i \)-th masked token, \( U \) the unmasked query tokens, and \( V \) the video features supporting accurate prediction. After the initial alignment phase, the QR head is frozen and \(\mathcal{L}_{qr}\) is excluded from training and inference.

\vspace{-0.4em}
\subsubsection*{\textbf{Geom-Regularization.}}
\label{geom}
The heuristic regularizor (\emph{i.e.} Eq.(\ref{vallina_reg})) in conventional DER aims to mitigate overconfidence by suppressing evidence, particularly for samples with high error. However, excessive suppression can lead to underconfidence due to non-adaptive suppression and sample imbalance. To be clear, we first consider the minus gradient of $\mathcal{L}^\mathrm{R}_i$ for $\Phi$ as follows:
\begin{equation}
\label{eq:reg_grad}
-\nabla_{\Phi}\mathcal{L}^\mathrm{R}_i = - \Delta,
\end{equation}
\begin{figure}[ht]
    \centering
    \includegraphics[width=0.47\textwidth]{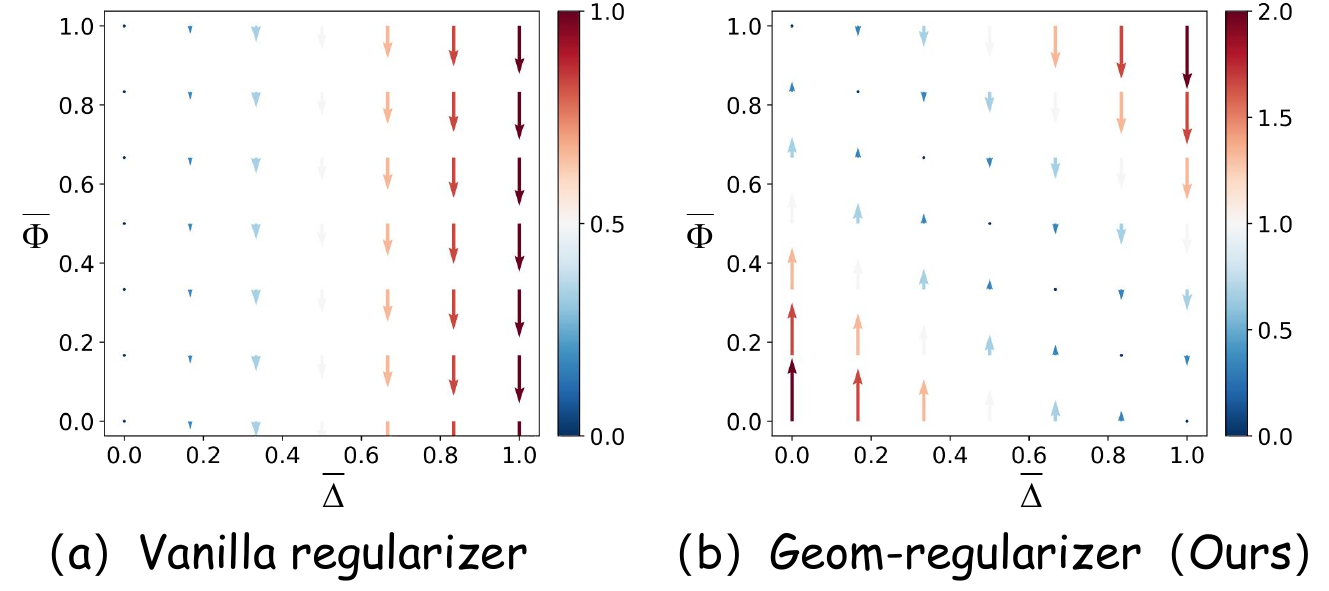}
    \vspace{-1.8em}
    \caption{Gradient field comparison. (a) Vanilla regularizer applies penalties based solely on error, decreasing evidence as error increases. (b) Our Geom-regularizer modulates penalties dynamically based on error magnitude and evidence levels. Our approach reflects the principle that accurate predictions should have higher evidence, while evidence should be suppressed for less accurate predictions.}
    \label{fig:gradient}
    \vspace{-1.6em}
\end{figure}
To explore the penalties bias in the vanilla regularizer, we visualized its optimization direction by examining the gradient field derived from the Eq.(\ref{eq:reg_grad}). As shown in Fig~\ref{fig:gradient} (a), the gradient is solely linked to the error and not to the evidence, indicating that the model cannot ascertain when the evidence has been adequately suppressed. This approach often results in insufficient gradients for batches dominated by small errors, potentially leading to biased penalties on evidence. 
As the model converges, the dominance of low error samples with small gradients skews the batch's average gradient. Consequently,  their evidence is over-suppressed, while high error samples see their evidence neglected or adversely adjusted, as shown in Figure~\ref{fig:evo}.

To overcome these limitations, we introduce Geom-regularization, inspired by~\cite{amini2020deep}, promoting the principle that ``\textbf{\textit{accurate predictions should have high evidence, while inaccurate ones should have low evidence}}". This approach provides more rational constraints rather than merely suppressing evidence. Initially, we normalize $\Delta$ to $\overline{\Delta}$ and $\Phi$ to $\overline{\Phi}$ (details in\textbf{ supplementary material}), which ensures that the model assigns $\overline{\Phi}=1$ to samples with $\overline{\Delta}=0$, and $\overline{\Phi}=0$ to samples with $\overline{\Delta}=1$.
We then ensure that the points $(\overline{\Delta},\overline{\Phi})$ closely follow the line $\overline{\Phi}+\overline{\Delta} = 1$ using a line regularizer as below:
\begin{equation}
\label{loss:type1}
\mathcal{L}^\text{L}_i(\boldsymbol{w})=\|\overline{\Phi}+\overline{\Delta}-1\|^2_2,
\end{equation}
we can follow the analysis for $\mathcal{L}^\mathrm{R}_i$. The minus gradient of $\mathcal{L}^\text{L}_i$ with respect to $\overline{\Phi}$ as below:
\begin{equation}
-\nabla_{\overline{\Phi}}\mathcal{L}^\text{L}_i = -2(\overline{\Delta}+\overline{\Phi}-1),
\end{equation}
which indicates this simple regularizer offers a gradient that relates to both error and evidence, enabling adaptive evidence suppression, as illustrated in Figure~\ref{fig:gradient} (b).


Our training objective for the evidential head is the combination of NLL and Geom-regularization:
\begin{equation}
\label{loss:der}
\mathcal{L}^\text{e}_i(\boldsymbol{w}) = 
\lambda_\text{NLL}  \mathcal{L}^{\mathrm{NLL}}_{i} +  \lambda_\text{geom} \mathcal{L}^\text{L}_i(\boldsymbol{w}), 
\end{equation}

To this end, our total loss can be formulated by a combination of MR loss $\mathcal{L}_\text{mr}$ and our evidential loss:
\begin{equation}
\label{loss:all}
\mathcal{L} =\mathcal{L}_\text{mr} + \lambda_\text{der} \frac{2}{N} \sum_{i=1}^N \mathcal{L}^\text{e}_i(\boldsymbol{w}) + \mathcal{L}_{qr},
\end{equation}
where $N$ symbolizes the number of clips in a training set.

%% file: Contents/Exp_new.tex
\section{Experiment}
\label{sec:exp}

\subsection{Datasets and Implementation Details}

\subsubsection*{\textbf{Datasets and Metrics.}} 
We evaluate on diverse public datasets: Charades-STA~\cite{gao2017tall} (indoor activities), QVHighlights~\cite{lei2021detecting} (untrimmed vlogs/news), and TACoS~\cite{regneri2013grounding} (cooking scenes). To assess robustness under temporal bias, we also use debiased ActivityNet-CD and Charades-CD~\cite{Lan2022ACL}. Dataset details and hyperparameters are provided in the \textbf{supplementary material}. And we report Recall@1 at IoU 0.5/0.7, mAP@0.75 and mAP avg (mean MAP over IoU 0.5-0.95, step 0.05), namely mAP.

\subsubsection*{\textbf{Experimental Settings.}} 
Following~\cite{lin2023univtg,li2024momentdiff}, we use CLIP~\cite{radford2021learning} (ViT-B/32) and SlowFast~\cite{feichtenhofer2019slowfast} (ResNet-50) as frozen backbones. The number of RFF blocks is set to 4. Training is two-stage: (1) QR masks/reconstructs 1 noun per sentence (default), using spaCy~\cite{spacy2} for noun extraction; QR runs for 30 epochs with a 1e-5 learning rate. (2) DEMR predicts bounding boxes per video clip; DER gradients in Eq.~(\ref{loss:type1}) are detached to focus on uncertainty optimization. NMS with threshold 0.7 is applied at evaluation. Unless stated, line regularizer is used on the evidential head. All experiments run on four Tesla V100 GPUs.
\vspace{-0.6em}

\subsection{Quantitative Results}
\begingroup
\setlength{\tabcolsep}{2pt} 
\renewcommand{\arraystretch}{1} 
\begin{table}[ht]
\caption{
Performance comparison with traditional ViT- and CNN-based methods on QVHighlights (\textit{val.} set), TACoS, and Charades-STA. \textbf{Bold}: best, \underline{underline}: second. This table's results are not based on any additional pre-training data.
}
\label{table:combined_single_narrow}
\vspace{-0.8em}
\centering
{\scriptsize 
\resizebox{\linewidth}{!}{
\begin{tabular}{c|ccc|ccc|ccc}
\hlineB{2.5}
\multirow{2}{*}{Method} & 
\multicolumn{3}{c|}{\cellcolor{yellow!30}QVHighlights} & 
\multicolumn{3}{c|}{\cellcolor{yellow!30}TACoS} & 
\multicolumn{3}{c}{\cellcolor{yellow!30}Charades-STA} \\
 & \cellcolor{yellow!30}R1@0.5 & \cellcolor{yellow!30}R1@0.7 & \cellcolor{yellow!30}mAP 
 & \cellcolor{yellow!30}R1@0.5 & \cellcolor{yellow!30}R1@0.7 & \cellcolor{yellow!30}mIoU 
 & \cellcolor{yellow!30}R1@0.5 & \cellcolor{yellow!30}R1@0.7 & \cellcolor{yellow!30}mIoU \\

 \hlineB{2.5}
M-DETR~\cite{NEURIPS2021_62e09734} & 53.9 & 34.8 & 30.7 & 28.0  & 12.9  & 27.2 & 46.0  & 27.5 & 41.3 \\
UMT~\cite{Liu_2022_CVPR} & 60.3 & 44.3 & 38.6 & 23.5  & 13.2  & 25.0 & 42.7  & 24.1 & 41.6 \\
QD-DETR~\cite{Moon_2023_CVPR}  & 62.7 & 46.7 & 41.2 & 24.7  & 12.0  & 25.5 & 52.1 & 30.6  & 45.5 \\
UniVTG~\cite{lin2023univtg} & 59.7 & - & 36.1 & 35.0  & 17.4  & 33.6 & \underline{58.0}  & \underline{35.7} & \underline{50.1} \\
EaTR~\cite{Jang_2023_ICCV} & 61.4 & 45.8 & 41.7 & - & - & - & - & - & - \\
CG-DETR~\cite{moon2023correlation} & \textbf{67.4} & \textbf{52.1} & 42.9 & \textbf{39.5}  & \textbf{23.4}  & \textbf{37.4} & 58.4  & 36.3 & 50.1 \\
MomentDiff~\cite{li2024momentdiff} & 57.4 & 39.7 & 36.0 & 33.7  & -  & - & 55.6  & 32.4 & - \\
\hline
\rowcolor{cyan!10}
\textbf{DEMR (Ours)} &\underline{65.0} & \underline{49.4} & \textbf{43.0} & \underline{37.3} & \underline{19.4} & \underline{33.9} & \textbf{60.2} & \textbf{38.0} & \textbf{51.6} \\ 
\hlineB{2.5}
\end{tabular}
}
}
\vspace{-1.2em}
\end{table}
\endgroup
\begin{table}[ht]
    \caption{Comparison on QVHighlights \textit{test} set with prevalent methods based on Multimodal Large Language Model (MLLM) backbone, ~\emph{i.e.} BLIP-2.}
    \vspace{-1em}
    \centering
    \scriptsize
    \renewcommand{\arraystretch}{1.0}
    \setlength{\tabcolsep}{5pt}
    \begin{tabular}{lcccc}
        \toprule
        \rowcolor{yellow!30}
        Method & R1@0.5 & R1@0.7 & mAP@0.75 & mAP \\
        \midrule
        Mr. BLIP~\cite{meinardus2024surprising} & 74.77 & 60.51 & 53.38 & 51.37 \\
        LLaVA-MR~\cite{lu2024llava} & \textbf{76.59} & \underline{61.48} & \underline{54.40} & -- \\
        \rowcolor{cyan!10}
        Ours & \underline{76.36} & \textbf{62.91} & \textbf{56.82} & \textbf{52.32} \\
        \bottomrule
    \end{tabular}
    \label{tab:qvhighlights_mllm}
    \vspace{-2em}
\end{table}

\begin{table}[ht]
    \caption{Ablation studies on the QVHighlights validation split. (a) Component Ablation: $\mathrm{Var}_{\mathrm{vis}}$ and $\mathrm{Var}_{\mathrm{text}}$ denote uncertainty variance as noise is added to visual or textual inputs, reflecting DEMR's sensitivity to each modality.}
    \centering
    \vspace{-1em}
    \scriptsize
    \renewcommand{\arraystretch}{1.0}
    \setlength{\tabcolsep}{10pt} 
    \begin{tabular}{l|cccc}
    \hlineB{2.5}
    \rowcolor{yellow!30}
    \multicolumn{1}{l|}{Method} & R1@0.5 & $\mathrm{Var}_{\mathrm{vis}}$ & $\mathrm{Var}_{\mathrm{text}}$ & $\Delta_{\mathrm{Var}}$ $\downarrow$\\
    \hlineB{2.5}
    Baseline & 61.1 & 9.17 & 0.85 & 8.32 \\
    \hspace{1em}+ RFF block & 62.4 & 8.63 & 1.60 & 7.03 \\
    \hspace{1em}+ QR & \underline{63.8} & 4.89 & 3.91 & \underline{0.98} \\
    \rowcolor{cyan!10}
    Full Model & \textbf{65.0} & 4.85 & 5.54 & \textbf{0.69} \\
    \hlineB{2.5}
    \end{tabular}
    \label{tab:performance_comparison_a}
    \vspace{-1em}
\end{table}

\subsubsection*{\textbf{Comparison with the state-of-the-art.}} For fair evaluation, we adopt the same backbone as most compared methods in both Table~\ref{table:combined_single_narrow} and Table ~\ref{tab:qvhighlights_mllm}. DEMR is benchmarked against leading traditional and MLLM-based approaches, demonstrating notable competitiveness across all datasets. Although this work primarily focuses on robust evidential learning for MR tasks, DEMR still achieves strong results without relying on dense clip-word guidance (CG-DETR) or dense frame sampling and temporal encoding (LLaVA-MR). This highlights the effectiveness of our approach and suggests substantial potential for further improvement when integrated with advanced MLLM-based backbones.

\begin{figure*}[h]
\centering
\subfigure[Performance on different $\lambda_{\text{geom}}$ and $\lambda_{\text{der}}$.]{\includegraphics[width=0.42\linewidth]{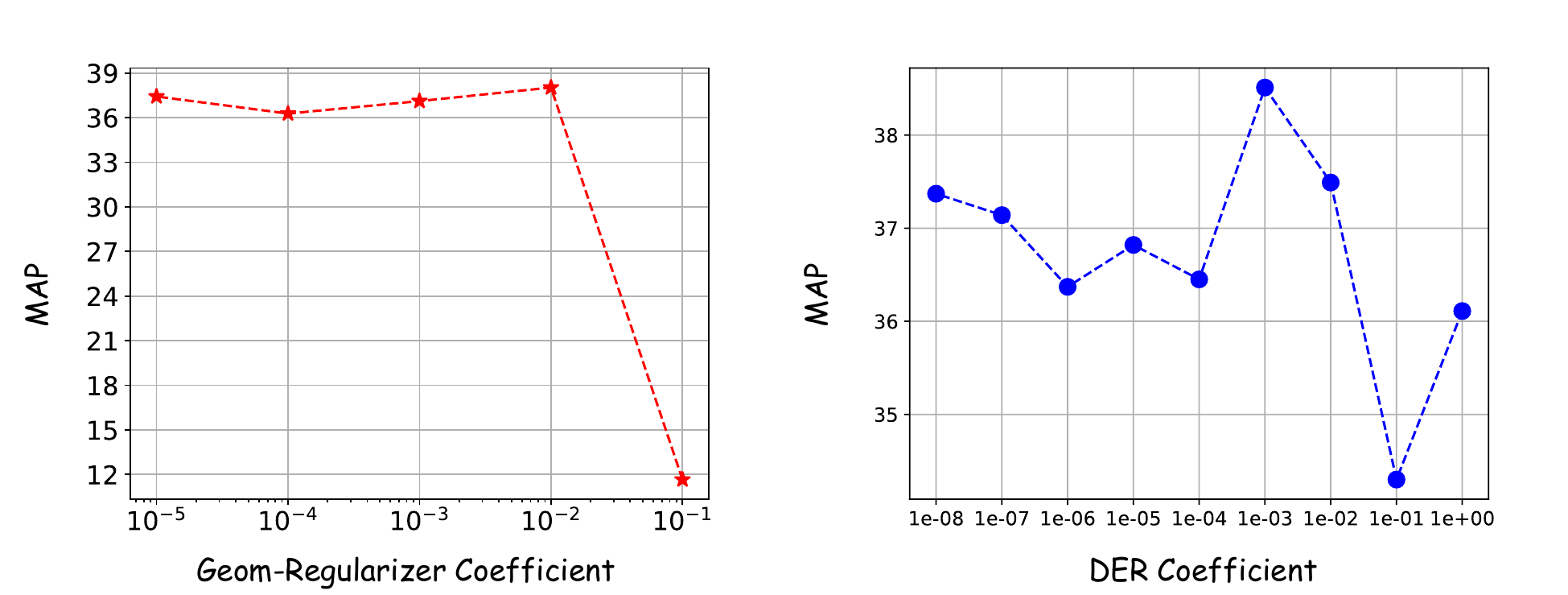}\label{fig:geom}}
\subfigure[Performance on different QR epochs and $\text{lr}$.]{\includegraphics[width=0.48\linewidth]{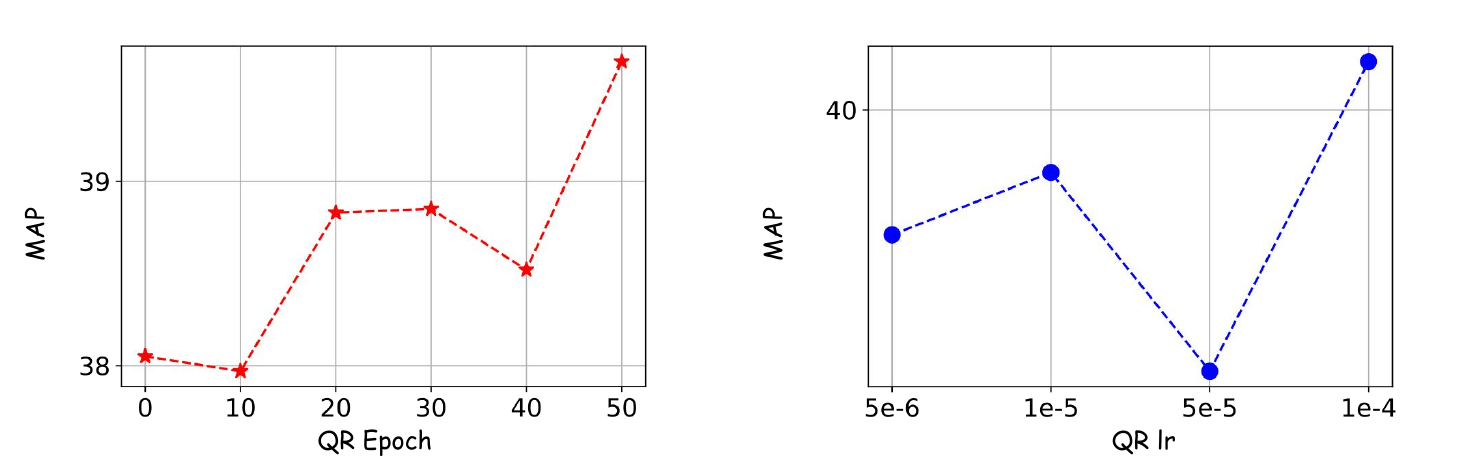}\label{fig:der}}
\vspace{-1.2em}
\caption{Parameters Analysis on QVHighlights val split. We examined the change of MAP. (a) Evaluate the effectiveness of our proposed Geom-regularizer (left) and der loss (right) under different weights. (b) Demonstrates the impact of the query reconstruction task at different epochs (left) and learning rates (right).}
\label{fig:ablation_mlm}
\vspace{-1em}
\end{figure*}

\vspace{-0.2cm}
\subsubsection*{\textbf{Ablation Study.}} 
\label{section:ablation}
\begin{figure}[ht]
    \centering
    \vspace{-0.4em}
    \includegraphics[width=0.48\textwidth]{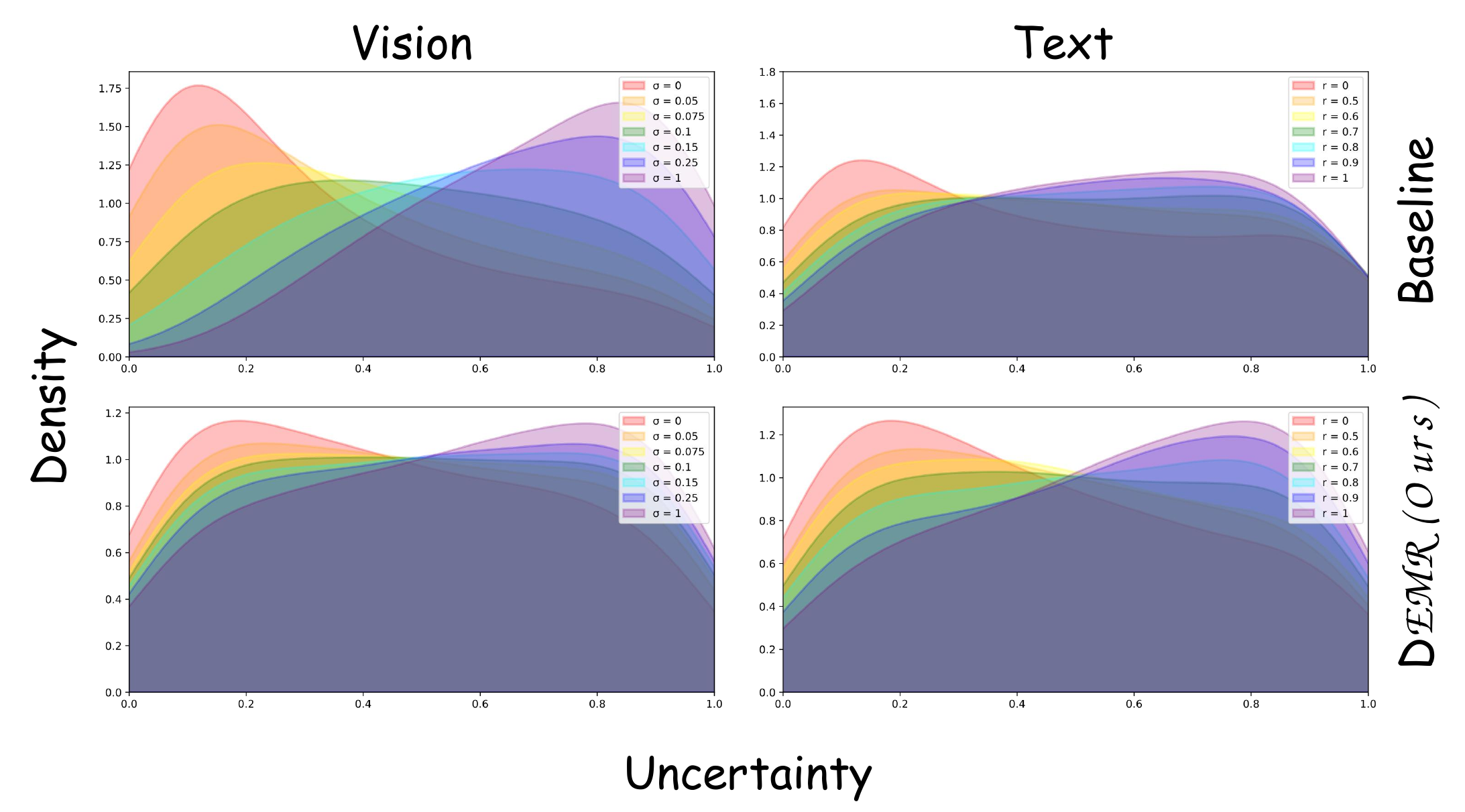}
    \vspace{-1.8em}
    \caption{\textbf{Uncertainty KDE over differect noise level}. Using Gaussian kernel density estimation (KDE), we plotted the uncertainty distribution for the QVHighlighte val set.}
    \label{fig:rainbow}
    \vspace{-1.4em}
\end{figure}

To evaluate the effectiveness of our debiasing method, we introduce targeted metrics to quantify modality imbalance on the QVHighlights validation set~\cite{lei2021detecting}. Specifically, we define $\mathrm{Var}{\mathrm{vis}}$, $\mathrm{Var}{\mathrm{text}}$, and $\Delta_{\mathrm{Var}}$ to measure uncertainty sensitivity under controlled noise. Gaussian noise is added to video embeddings, and a proportion of text tokens is replaced with irrelevant content, following the schedule in Fig.~\ref{fig:rainbow}. We compute the average variance of uncertainty for each modality and use $\Delta_{\mathrm{Var}}$ to assess balance. As shown in Table~\ref{tab:performance_comparison_a}(a), both the RFF block and QR task substantially reduce modality bias, confirming the effectiveness of our approach.
\vspace{-0.6em}

\subsection*{Parameter Analysis}
\textbf{Module Hyperparameters.}  
We analyze the sensitivity of MAP to key hyperparameters (Figure~\ref{fig:geom}). MAP peaks at \(\lambda_{\text{geom}} = 10^{-2}\), which we adopt. For \(\lambda_{\text{der}}\), performance remains stable at low values but degrades when \(\lambda_{\text{der}}\) exceeds \(1 \times 10^{-2}\), indicating that excessive uncertainty constraints harm grounding. Thus, we set \(\lambda_{\text{der}} = 1 \times 10^{-3}\). Figure~\ref{fig:ablation_mlm}(b) shows that increasing QR epochs from 0 to 50 significantly boosts MAP (+1.5\%), with optimal QR learning rate at \(1 \times 10^{-4}\). Full details are in the \textbf{Supplementary Material}.
\textbf{Auxiliary QR Task.}  
Table~\ref{tab:qr_results} demonstrates that the QR task substantially improves text-video alignment. High mask ratios reduce context and hurt performance, especially when all nouns, which are key semantic carriers, are masked. Masking a single noun preserves context and optimizes alignment. Careful mask ratio selection, particularly for nouns, is thus critical for maximizing QR benefits.
\vspace{-0.7em}
\subsection{Qualitive Resluts}
\subsubsection*{\textbf{Uncertainty Sensitivity to Modalities.}}
To more clearly demonstrate the role of the proposed RFF block and QR task in promoting modality balance and alignment, we conduct fair adversarial experiments on the QVHighlights validation set to compare the modality sensitivity of the baseline model and DEMR. We apply the noise addition method mentioned in ablation and used the noise intensity format shown in Figure~\ref{fig:rainbow}. Also as illustrated in Figure~\ref{fig:rainbow}, when progressively increasing noise levels for two different modalities, the output uncertainty of the baseline model (Top row) shows high sensitivity to the visual modality and low sensitivity to the textual modality. In contrast, the DEMR achieves balanced uncertainty across both modalities. Regardless of which modality the noise is added to, the uncertainty distribution of the model shifts from left-skewed to right-skewed at the same rate.
\vspace{-0.6em}
\begin{figure}[ht]
\vspace{-0.3cm}
\centering
\includegraphics[width=1\linewidth]{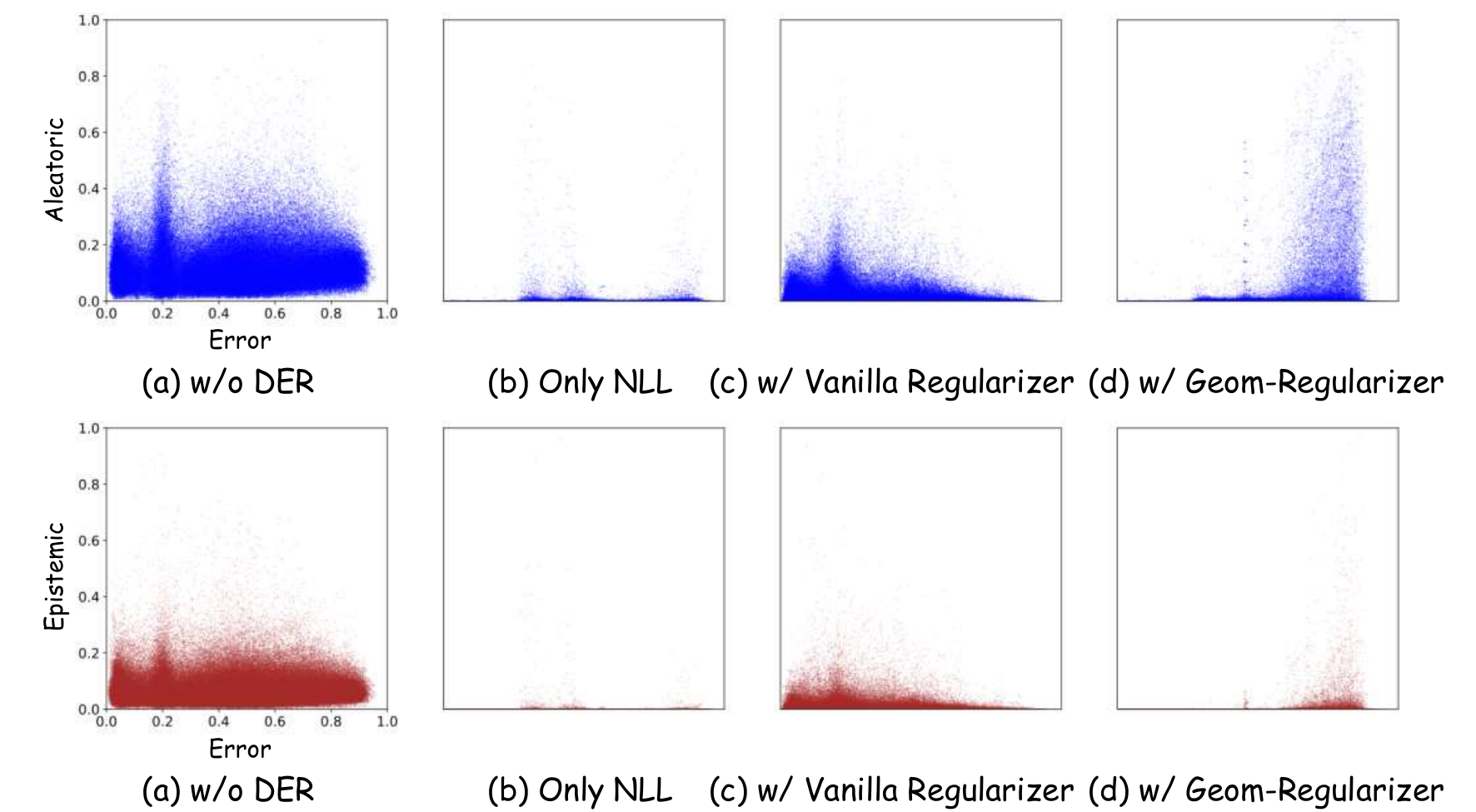}
\vspace{-2em}
\caption{Effects of Various Regularization Techniques on Uncertainty Distribution. (a)-(d) illustrate the impact of different regularization methods on the relationship between aleatoric uncertainty (top row) and epistemic uncertainty (bottom row) with respect to prediction error. The models include: (a) without DER, (b) only NLL, (c) with Vanilla Regularizer, and (d) with Geom-Regularizer.}
\label{fig:calibration}
\vspace{-1.4em}
\end{figure}
\begin{figure}[ht]
\centering
\includegraphics[width=1\linewidth]{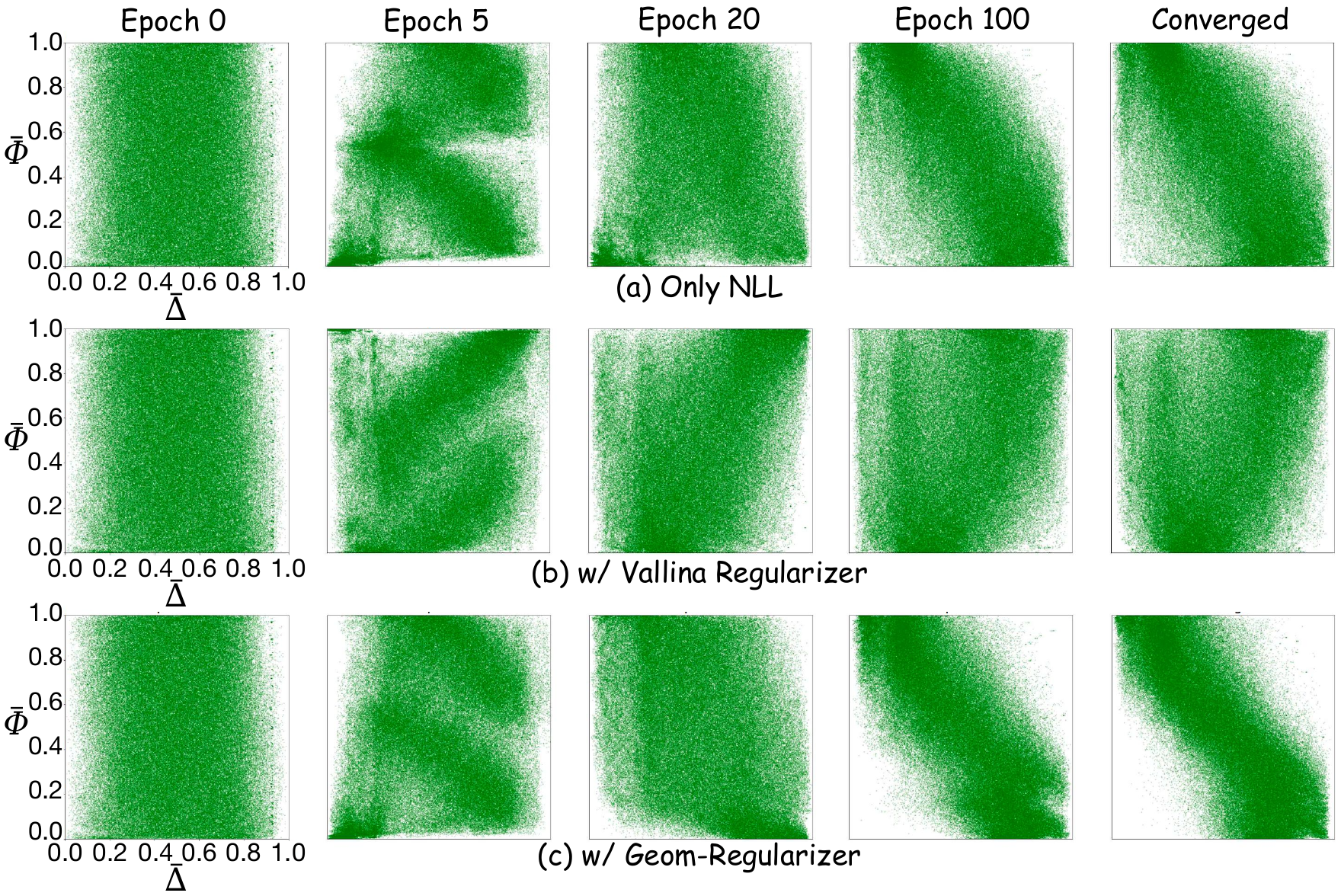}
\vspace{-2em}
\caption{Illustration of the different regularizer impact. As the training progresses, the model's uncertainty is optimized in the right direction with our regularizer.}
\label{fig:evo}
\end{figure}

\begin{table}[ht]
    \centering
    \scriptsize
    \setlength{\tabcolsep}{5pt}
    \caption{Impact of different mask ratio on task performance.}
    \vspace{-1.5em}
    \begin{tabular}{lcccccc}
        \toprule
        \rowcolor{yellow!30}
        Mask Ratio & 0\% & 1 noun & 25\% & 50\% & 75\% & all noun \\
        \midrule
        MR-full-mAP-key & 32.75 & \textbf{36.93} & 33.40 & 32.43 & 31.80 & 31.71 \\
        MR-full-R1@0.5-key & 57.23 & \textbf{64.06} & 61.74 & 59.55 & 59.23 & 58.52 \\
        MR-full-R1@0.7-key & 37.11 & \textbf{43.61} & 39.17 & 37.13 & 34.26 & 34.45 \\
        \bottomrule
    \end{tabular}
    \label{tab:qr_results}
    \vspace{-1.6em}
\end{table}

\subsubsection*{\textbf{Uncertainty Calibration.}}
\label{scatter}
Since we propose geom-regularizor to calibrate of counterintuitive uncertainty prediction, we aim to assess the efficacy of our approach by contrasting the performance of aleatoric and epistemic uncertainty estimation with and without our regularizer, as well as against using vanilla regularization in~\cite{amini2020deep}. Ideally, optimal uncertainty measures should effectively identify deviations in predictions (\emph{i.e.}, take high uncertainty when the model is making errors). Figure~\ref{fig:calibration} illustrates our comparison of different regularization methods on the QVHighlights validation set. The horizontal axis of each scatter plot represents $\overline{\Delta}$ (\emph{i.e.} normalized error), while the vertical axis represents one of the two types of uncertainty. It indicates that without DER, the model fails to manage uncertainty, resulting in unreliable inferences. Using only NLL leads to overconfidence, as the model exhibits low uncertainty across all error levels, suggesting overfitting. The Vanilla Regularizer presents a paradox, showing higher uncertainty at lower error rates, which is counterintuitive. In contrast, our Geom-Regularizer effectively calibrates uncertainty, indicating higher uncertainty at greater error rates and reducing overconfidence.

\begin{figure}[ht]
\centering
\vspace{-1em}
\includegraphics[width=1\linewidth]{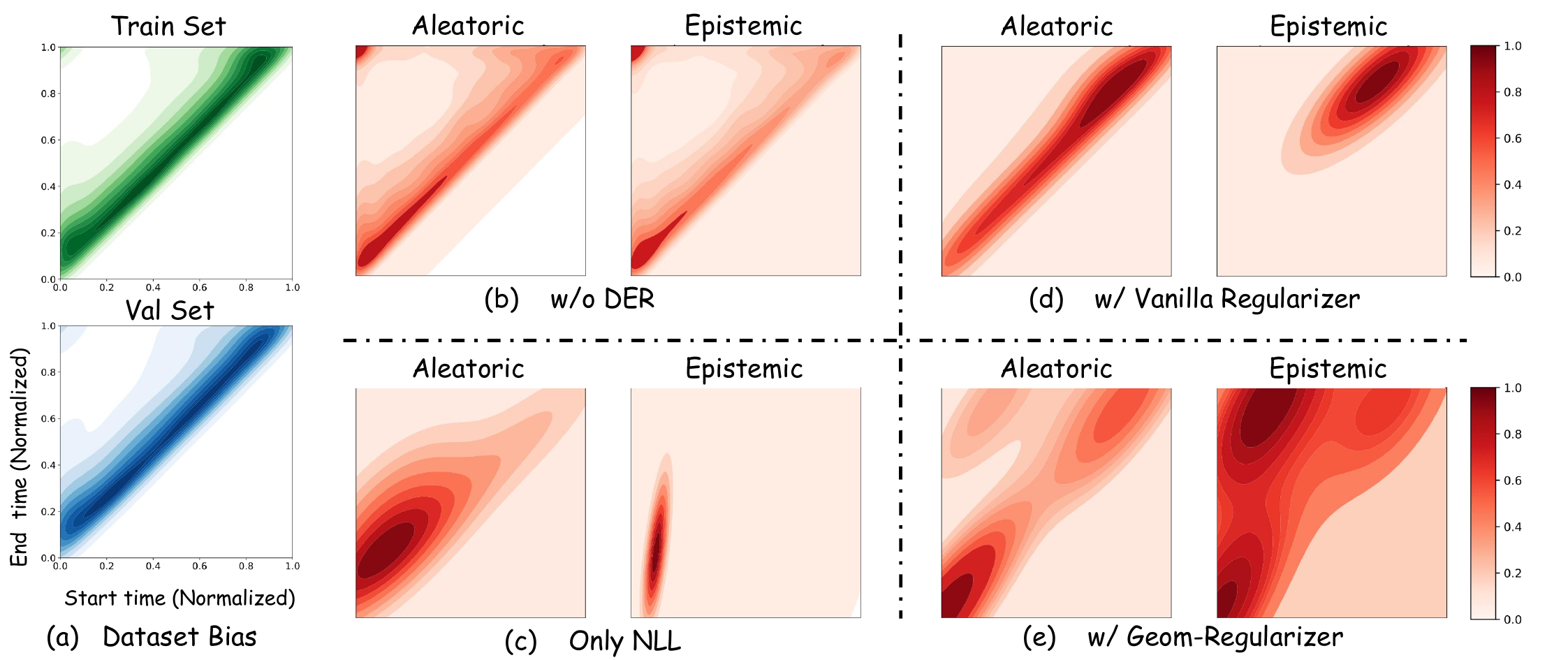}
\vspace{-2em}
\caption{\textbf{Dataset bias sensitivity.} (a) Joint distributions of the start and end timestamps of the ground-truth moments in the QVHighlights dataset. (b), (c), (d), and (e) show the predicted uncertainty's sensitivity to temporal biases in the dataset under different conditions.}
\label{fig:bias_qv}
\vspace{-1.8em}
\end{figure}
As illustrated in Figure~\ref{fig:evo}, it is obvious that ``\textbf{\textit{accurate predictions with high evidence while inaccurate predictions with low evidence}}" has been reflected in the knowledge of model with only NLL. Unfortunately, the vanilla regularizer excessively suppress the evidence of low error predictions, but ignores and even enlarges the evidence of high error predictions. Geom-regularizer turn the situation around, retain the main knowledge learned by NLL, and provides calibration for more reasonable uncertainty estimation.

\subsubsection*{\textbf{Temporal Bias Sensitivity.}}
Most moment retrieval datasets exhibit significant temporal bias in moment duration and position, leading to underrepresented (Temporal OOD) regions, as visualized in Figure~\ref{fig:bias_qv}(a) using QVHighlights. Higher epistemic uncertainty is expected in these OOD regions. We evaluate model uncertainty across temporal regions under different settings. Without DER constraints, the evidential head's uncertainty merely reflects the biased data distribution (Figure~\ref{fig:bias_qv}(b)). Using only NLL loss, the model shows low epistemic uncertainty, indicating overconfidence (Figure~\ref{fig:bias_qv}(c)). Vanilla regularization reduces concentrated uncertainty but lacks OOD sensitivity (Figure~\ref{fig:bias_qv}(d)). In contrast, our Geom-regularizor significantly increases epistemic uncertainty in OOD regions (Figure~\ref{fig:bias_qv}(e)), enhancing temporal bias awareness.
Table~\ref{tab:temp_bias} further shows that DEMR achieves robust cross-distribution performance, notably surpassing MomentDiff and MomentDETR by 3.37\% and 10.47\% respectively on Charades-CD R1@0.3. Importantly, DEMR narrows the IID-OOD gap to 3.29\%, compared to 12.00\% for CM-NAT, demonstrating improved generalization and reduced sensitivity to dataset bias. These results confirm DEMR's effectiveness in addressing temporal bias and enhancing adaptability in challenging retrieval scenarios.

\begin{table}[ht]
    \centering
    \scriptsize
    \setlength{\tabcolsep}{3pt}
    \renewcommand{\arraystretch}{0.9}
    \caption{Performance comparison on Charades-CD and ActivityNet-CD datasets. N/A denotes that the data have not been open-source and thus not accessible.}
    \vspace{-1.5em}
    \begin{adjustbox}{width=\linewidth}
    \begin{tabular}{l|ccc|ccc}
        \toprule
        \rowcolor{yellow!30}
        \textbf{Method} & \textbf{R1@0.3} & \textbf{R1@0.5} & \textbf{R1@0.7} & \textbf{R1@0.3} & \textbf{R1@0.5} & \textbf{R1@0.7} \\
        \midrule
        DRN~\cite{zeng2020dense} (i.i.d) $\uparrow$ & 51.35 & 41.91 & 26.74 & 48.92 & 39.27 & 25.71 \\
        DRN~\cite{zeng2020dense} (o.o.d) $\uparrow$ & 40.45 & 30.43 & 15.91 & 36.86 & 25.15 & 14.33 \\
        $\Delta$ ↓ & 10.90 & 11.48 & 10.83 & 12.06 & 14.12 & 11.38 \\
        \midrule
        TSP-PRL~\cite{Wu2020TreeStructuredPB} (i.i.d) $\uparrow$ & 46.44 & 35.43 & 17.01 & 44.93 & 33.93 & 19.50 \\
        TSP-PRL~\cite{Wu2020TreeStructuredPB} (o.o.d) $\uparrow$ & 31.93 & 19.37 & 6.20 & 29.61 & 16.63 & 7.43 \\
        $\Delta$ ↓ & 14.51 & 16.06 & 10.81 & 15.32 & 17.30 & 12.07 \\
        \midrule
        2D-TAN~\cite{Zhang2019Learning2T} (i.i.d) $\uparrow$ & 53.71 & 46.48 & 28.18 & 49.18 & 40.87 & 28.95 \\
        2D-TAN~\cite{Zhang2019Learning2T} (o.o.d) $\uparrow$ & 43.45 & 28.18 & 13.73 & 30.86 & 18.86 & 9.77 \\
        $\Delta$ ↓ & 10.26 & 18.30 & 14.45 & 18.32 & 22.01 & 19.18 \\
        \midrule
        MMN~\cite{Wang2022MMN} (i.i.d) $\uparrow$ & N/A & N/A & N/A & N/A & N/A & N/A \\
        MMN~\cite{Wang2022MMN} (o.o.d) $\uparrow$ & 55.91 & 34.56 & 15.84 & 44.13 & 24.69 & 12.22 \\
        $\Delta$ ↓ & N/A & N/A & N/A & N/A & N/A & N/A \\
        \midrule
        CM-NA~\cite{Lan2023CurriculumMA} (i.i.d) $\uparrow$ & 64.21 & 53.82 & 34.47 & 49.91 & 41.67 & 28.82 \\
        CM-NA~\cite{Lan2023CurriculumMA} (o.o.d) $\uparrow$ & 52.21 & 39.86 & 21.38 & 32.32 & 20.78 & 11.03 \\
        $\Delta$ ↓ & 12.00 & 13.96 & 13.09 & 17.59 & 20.89 & 17.79 \\
        \midrule
        MomentDETR~\cite{lei2021detecting_Moment-DETR} (i.i.d) $\uparrow$ & N/A & N/A & N/A & N/A & N/A & N/A\\
        MomentDETR~\cite{lei2021detecting_Moment-DETR} (o.o.d) $\uparrow$ & 57.34 & 41.18 & 19.31 & 39.98 & 21.30 & 10.58 \\
        $\Delta$ ↓ & N/A & N/A & N/A & N/A & N/A & N/A \\
        \midrule
        MomentDiff~\cite{li2024momentdiff} (i.i.d) $\uparrow$ & N/A & N/A & N/A & N/A & N/A & N/A \\
        MomentDiff~\cite{li2024momentdiff} (o.o.d) $\uparrow$ & 67.73 & 47.17 & 22.98 & 45.54 & 26.96 & 13.69 \\
        $\Delta$ ↓ & N/A & N/A & N/A & N/A & N/A & N/A \\
        \midrule
        \rowcolor{cyan!20}
        Ours (i.i.d) $\uparrow$ & 71.10 & 62.20 & 43.29 & 56.33 & 41.77 & 27.47 \\
        \rowcolor{cyan!20}
        Ours (o.o.d) $\uparrow$ & 67.81 & 52.46 & 30.97 & 41.64 & 23.76 & 16.89 \\
        \rowcolor{cyan!20}
        $\Delta$ ↓ & 3.29 & 9.74 & 12.32 & 14.69 & 18.01 & 10.58 \\
        \bottomrule
    \end{tabular}
    \end{adjustbox}
    \label{tab:temp_bias}
    \vspace{-0.6em}
\end{table}

\begin{figure}[h]
\centering
\vspace{-0.8em}
\includegraphics[width=\linewidth]{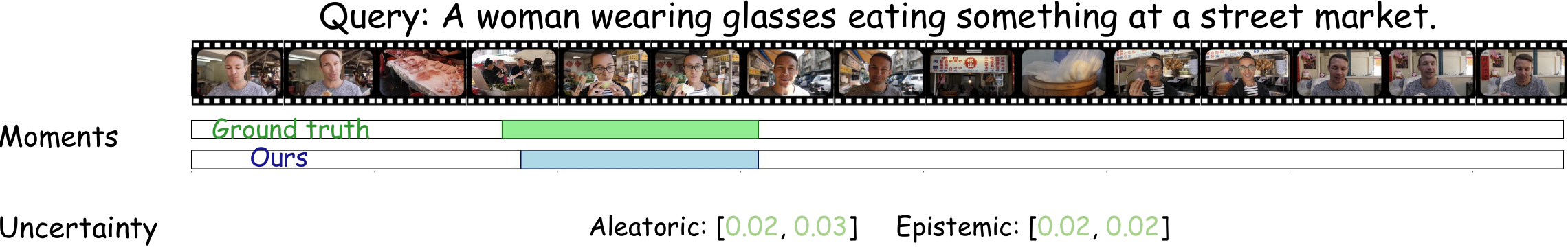}
\vspace{0.4em}
\includegraphics[width=\linewidth]{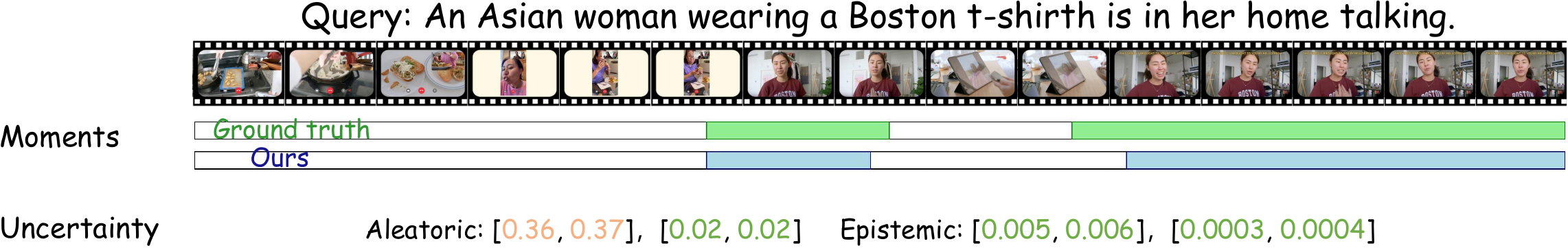}
\vspace{0.4em}
\includegraphics[width=\linewidth]{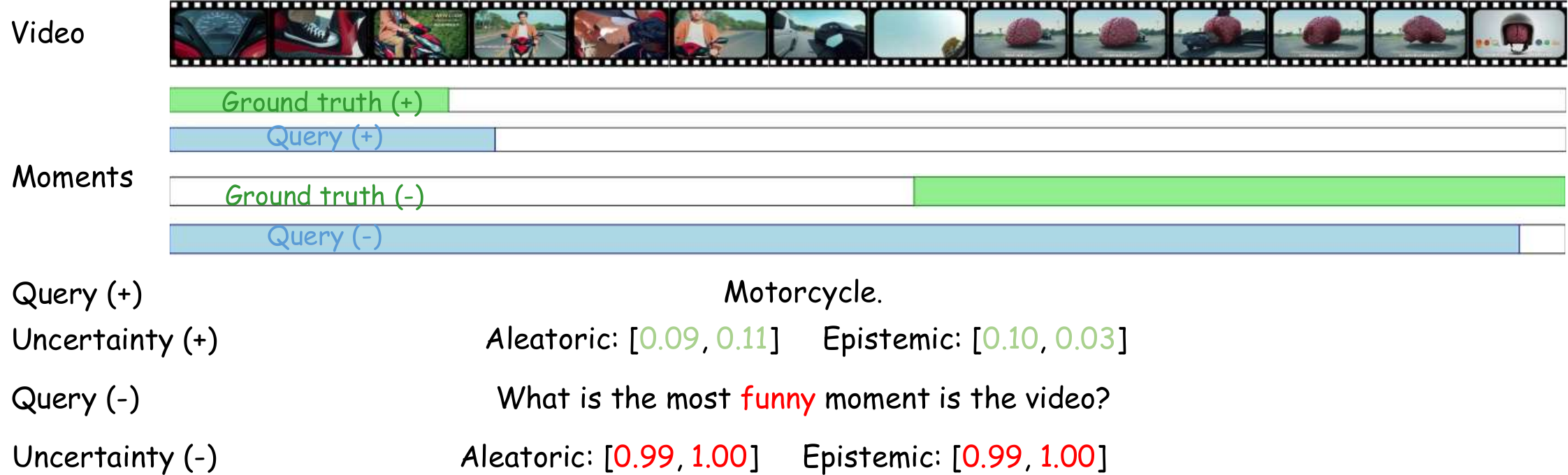}
\vspace{-2em}
\caption{Cases Study.}
\label{fig:cases_study}
\vspace{-1.6em}
\end{figure}

\subsubsection*{\textbf{Cases Study.}}
\label{sec:cases_study}
In Figure~\ref{fig:cases_study}, we showcase DEMR's effectiveness in complex scenarios. For ``a woman wearing glasses eating at a street market,'' DEMR accurately identifies key moments with low aleatoric and epistemic uncertainty, indicating stable and confident localization. In ``an Asian woman wearing a Boston t-shirt at home,'' DEMR remains consistent with ground truth despite higher aleatoric uncertainty from environmental variability, while epistemic uncertainty stays low. Furthermore, the bottom case demonstrates DEMR's uncertainty estimation on advertisement video with ambiguous queries. Although lacking humor-related annotations leads to biased results, DEMR reliably estimates uncertainty, supporting model adaptation to challenging samples during training.
\vspace{-1.4em}

%% file: Contents/Conclusion.tex
\section{Conclusion}
As AGI advances, MR models face increasing challenges from open-ended user inputs. In this paper, we propose DEMR, a robust MR model that addresses key biases in baseline evidential-based methods and enables effective uncertainty quantification, thus improving response credibility for hard cases. While current performance is constrained by data quality and scale, DEMR provides valuable strategies for enhancing the trustworthiness of AI decisions. Future work will further integrate DEMR with advanced MLLMs to expand its application and improve reliability in video tasks.